\documentclass{article} 

\usepackage[dvipsnames]{xcolor}

\usepackage{collas2024_conference,times}
\usepackage{easyReview}

\usepackage{amsmath,amsfonts,bm}

\def\eqref#1{equation~\ref{#1}}

\def\1{\bm{1}}

\DeclareMathAlphabet{\mathsfit}{\encodingdefault}{\sfdefault}{m}{sl}
\SetMathAlphabet{\mathsfit}{bold}{\encodingdefault}{\sfdefault}{bx}{n}

\usepackage{hyperref}
\hypersetup{
    colorlinks=true,
    linkcolor=red,
    filecolor=magenta,
    urlcolor=blue,
    citecolor=purple,
    pdftitle={Overleaf Example},
    pdfpagemode=FullScreen,
    }

\usepackage{subfigure}
\usepackage{arydshln}
\usepackage{algorithm}
\usepackage[algo2e,ruled,linesnumbered,lined,noend]{algorithm2e}
\usepackage[dvipsnames]{xcolor}
\usepackage{wrapfig}
\usepackage{booktabs}

\title{Channel-Selective Normalization for Label-Shift Robust Test-Time Adaptation}

\author{\parbox{\textwidth}{\centering
    Pedro Vianna$^{1,2,3}$
    \qquad Muawiz Chaudhary$^{4,5}$ \hspace{-10pt}
    \qquad Paria Mehrbod$^{4,5}$ \hspace{-10pt}
    \qquad An Tang$^{2,3}$\\
    Guy Cloutier$^{1,2,3}$
    \qquad Guy Wolf$^{3,4}$
    \qquad Michael Eickenberg$^{6}$
    \qquad Eugene Belilovsky$^{4,5}$} \vspace{5pt}\\
$^1$ Laboratory of Biorheology and Medical Ultrasonics; 
$^2$ University of Montreal Hospital Research Center;\\
$^3$ University of Montreal;
$^4$ Mila -- Quebec AI Institute;
$^5$ Concordia University;
$^6$ Flatiron Institute }

\preprintcopy 

\begin{document}

\maketitle

\begin{abstract}
Deep neural networks have useful applications in many different tasks, however their performance can be severely affected by changes in the 
input distribution. For example, in the biomedical field, their performance can be affected by changes in the data (different machines, populations) between training and test datasets. To ensure robustness and generalization to real-world scenarios, test-time adaptation has been recently studied as an approach to adjust models to a new data distribution during inference. Test-time batch normalization is a simple and popular method that achieved compelling performance on domain shift benchmarks. It is implemented by recalculating batch normalization statistics on test batches. Although initial work focused on analysis with test data that has the same label distribution as the training data, in many practical applications this technique is vulnerable to label distribution shifts, sometimes 
causing catastrophic failure. This presents a risk in applying test time adaptation methods in deployment. We propose to tackle this challenge by only selectively adapting channels in a deep network, minimizing drastic adaptation that is sensitive to label shifts. Our selection scheme is based on two principles that we empirically motivate: (1) later layers of networks are more sensitive to label shift; (2) individual features can be sensitive to specific classes. We apply the proposed technique to three classification tasks, including CIFAR10-C, Imagenet-C, and diagnosis of fatty liver, where we explore both covariate and label distribution shifts. We find that our method allows to bring the benefits of TTA while significantly reducing the risk of failure common 
in imbalanced scenarios.

\end{abstract}

\section{Introduction}
\label{sec:intro}

Deep learning models have shown excellent performance on a wide variety of tasks in computer vision research, such as image classification \citep{aleximagenet, yuexternalval}, segmentation \citep{wasserthalsegment, dingsegment}, and object detection \citep{Chendetect, Yindetect}. However, a commonly cited limitation of these models is the inability to generalize across different domains \citep{hendrycks2018}. Generalization can be simply defined as the ability of a deep learning algorithm, trained on one or more source domains, to be applied to a different yet still related target domain. Typically, in real-world deployment scenarios models might encounter data with critical differences, hampering their performance. This decrease in performance has been observed in multiple areas, including life-threatening contexts, such as autonomous driving \citep{kimmemory, vibashan} and medical diagnostics \citep{blaivas, wang}.  

Domain adaptation techniques aim to enhance a model’s generalizability, anticipating the possible challenges in the event of distribution shifts. Traditionally, domain adaptation assumes that the model has access to unlabeled test data and source data during the training, which may not be practical in some cases \citep{PivaWACV, liangSHOT}. A recently emerging technique to deal with distribution shift is test-time adaptation (TTA) \citep{sunTTT, liang2023ttasurvey}, a type of unsupervised domain adaptation, where unlabeled test data is used to update the model parameters at test-time, before predictions. It is often assumed that data arrives in batches, and some studies have proposed a setting of test-time batch adaptation that take advantage of batch-level information to adapt to the distribution shift \citep{TENT, DBLP:journals/corr/abs-2006-10963, chen2022contrastive}. 

Test-time batch normalization (TTN) \citep{DBLP:journals/corr/abs-2006-10963, DBLP:conf/nips/SchneiderRE0BB20} is an approach which replaces batch normalization statistics estimated as running averages on the training set with the statistics of the test data batch. Despite being a simple approach, it has been shown to improve robustness under covariate shift, handling particularly well various cases of image corruptions. Based on that, other TTA approaches apply TTN as a critical component in their foundation \citep{TENT, limTTN,delta}.  However TTN was proposed and  developed in the context of covariate shifts only, while in many realistic scenarios, such as medical imaging, the label distribution of data can shift from training to testing. 

In order to be practically useful, TTA methods relying on batch data need to be able to operate outside of this optimistic scenario of class-balanced training and testing settings, while simultaneously dealing with covariate shifts. As noted in recent work \citep{lame} label distribution shifts can cause failures. We emphasize that these failures can be catastrophic, thus creating risk in the usage of TTA when the label distribution under deployment 
is unknown. We observe that certain channels and layers can be much more sensitive with respect to label distribution shifts under TTN. Motivated by this, we propose a method to mitigate the risk of label distribution shift catastrophically affecting model predictions. 
Our method, based on TTN, aims to adapt only those channels of the batch normalization layers whose activation statistics do not exhibit strong correlation to label distribution. The proposed method is applied to classification of two well-known benchmark natural image datasets --- CIFAR-10 \citep{cifar} and ImageNet-1K \citep{imagenet} --- and on adaptation from a private medical dataset \citep{vianna} to a publicly available one \citep{byra}. When deployed in target data with different distribution, our proposed method is effective in obtaining the benefits of TTA while avoiding catastrophic failure cases, greatly improving the practical utility of TTN methods. 

Our contributions are summarized below:
\begin{enumerate}

    \item We propose a novel method that deals with label distribution shifts by leveraging information that can be easily computed at the end of model training and used during inference time to select channels sensitive to class information. We augment this channel selection strategy with a prior on the effect of depth on sensitivity to class information. 
    \item Experiments on three different datasets demonstrate that our method can be applied on a range of classification tasks, in particular for highly imbalanced target data. Notably, the adaptation across liver ultrasound datasets from different sites is shown to be effective. Our method is more robust than benchmark techniques across various degrees of label imbalance, showing similar performance in balanced cases and improvements in imbalanced cases. 
\end{enumerate}

\section{Related Work}
\label{sec:relwork}

Batch normalization \citep{DBLP:conf/icml/IoffeS15}, henceforth BatchNorm, is a technique that aims to stabilize the distributions of layer inputs by introducing additional network layers that control the first two moments (mean and variance) of these distributions. This positively affects the optimization process of deep neural networks during training, reduces overfitting, and prevents the exploding or the vanishing gradient problems.

Even though BatchNorm is effective, widely accepted, and commonly used in many deep neural networks, its mechanisms are not entirely understood. Methods extending this approach provide alternatives to BatchNorm, such as layer normalization \citep{DBLP:journals/corr/BaKH16}, instance normalization \citep{DBLP:journals/corr/UlyanovVL16}, and normalization based on groups of channels \citep{DBLP:conf/eccv/WuH18}. Furthermore, weight normalization \citep{DBLP:conf/nips/SalimansK16} extends this line of work, modifying the weights of the models instead of features.

Various studies have been conducted to further understand how batch-normalization works. The original motivation for BatchNorm refers to the problem of internal covariate shift, however a later study \citep{DBLP:conf/nips/SanturkarTIM18} notes that it is not an appropriate explanation. Rather, the loss landscapes of models with BatchNorm appear to be smoother than those of equivalent models without it, allowing for use of larger range of learning rates and faster network convergence.  Moreover, by investigating randomly initialized networks with all layers frozen and adapting only the BatchNorm layers, it has been shown that BatchNorm layers have substantial expressive power, achieving surprisingly good results on CIFAR-10 and ImageNet \citep{DBLP:conf/iclr/FrankleSM21}.

Test-time adaptation methods often involve updating the parameters associated with the BatchNorm layers to distributions with covariate shift \citep{DBLP:conf/nips/SchneiderRE0BB20, DBLP:journals/corr/abs-2006-10963}. This approach is closely related to the unsupervised domain adaptation method AdaBN \citep{li2016revisiting, DBLP:journals/corr/abs-2103-05898}, and has achieved promising results in mitigating image corruptions. Recent TTA methods have been published using entropy minimization on the BatchNorm affine parameters \citep{TENT}, but they appear to be biased towards the dominant classes \citep{park2023label}. In contrast, a distinct class of methods focuses on TTA without relying on BatchNorm adjustments, proposing techniques that leverage ideas from self-supervised learning \citep{sun2020test, chen2022contrastive}, test-time augmentations \citep{mixnorm, sita}, or sample selection \citep{eata}. 

A common limitation in TTA studies is regarding label distribution shifts, as most of the proposed methods fail to consider the varying class balance in the test sets, eventually falling into the ``winner-takes-all'' dilemma in which the majority classes dominate the predictions \citep{you2021, liang2023ttasurvey}. To address the class-imbalanced domain adaptation challenge, some works focus on estimating the test set distribution \citep{sun2022, lipton2018, park2023label}.

Recently, other works have also proposed methods for dealing with label distribution shift in TTA. LAME \citep{lame} performs the online adaptation of the classification outputs instead of the model's parameters, making it agnostic to training and testing conditions, but it does not perform well in balanced scenarios. DELTA \citep{delta} uses test-time batch renormalization and dynamic online reweighting in an online manner to avoid mini-batch bias. A label shift adapter was designed \citep{park2023label} to improve TTA performance in scenarios where both source and target are imbalanced, by estimating the distribution of the target set. However, it uses a large number of hyperparameters and is specialized to the case of imbalanced source data. 
Focusing on learned features, in few-shot learning it was observed that trained models presents channel bias, and a channel-wise transformation during test-time can improve predictions in out-of-distribution tasks \citep{luo_fsl}.

\section{Hybrid Test-Time Batch Normalization}
\label{sec:method}

\begin{figure}
\centering
\begin{subfigure} \centering
    \includegraphics[width=0.45\textwidth]{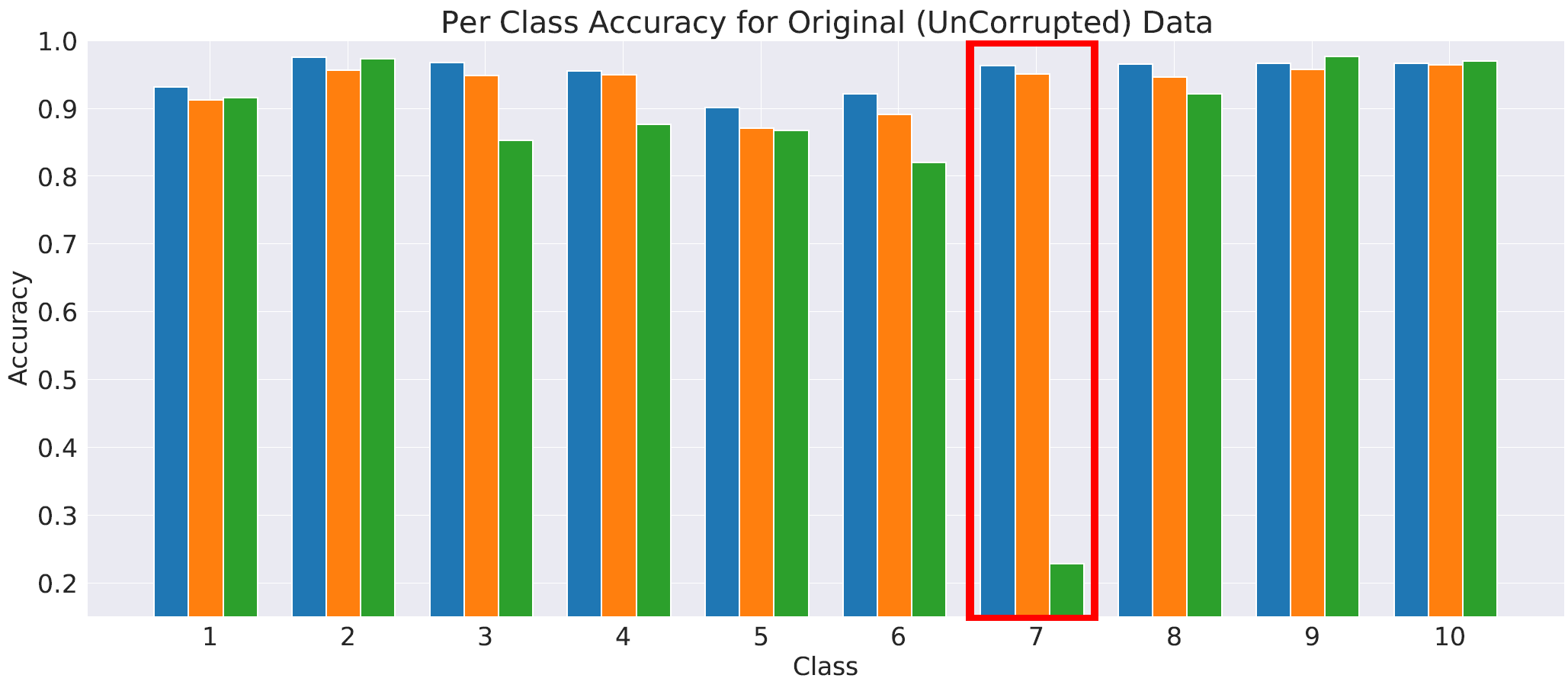}
    \includegraphics[width=0.45\textwidth]{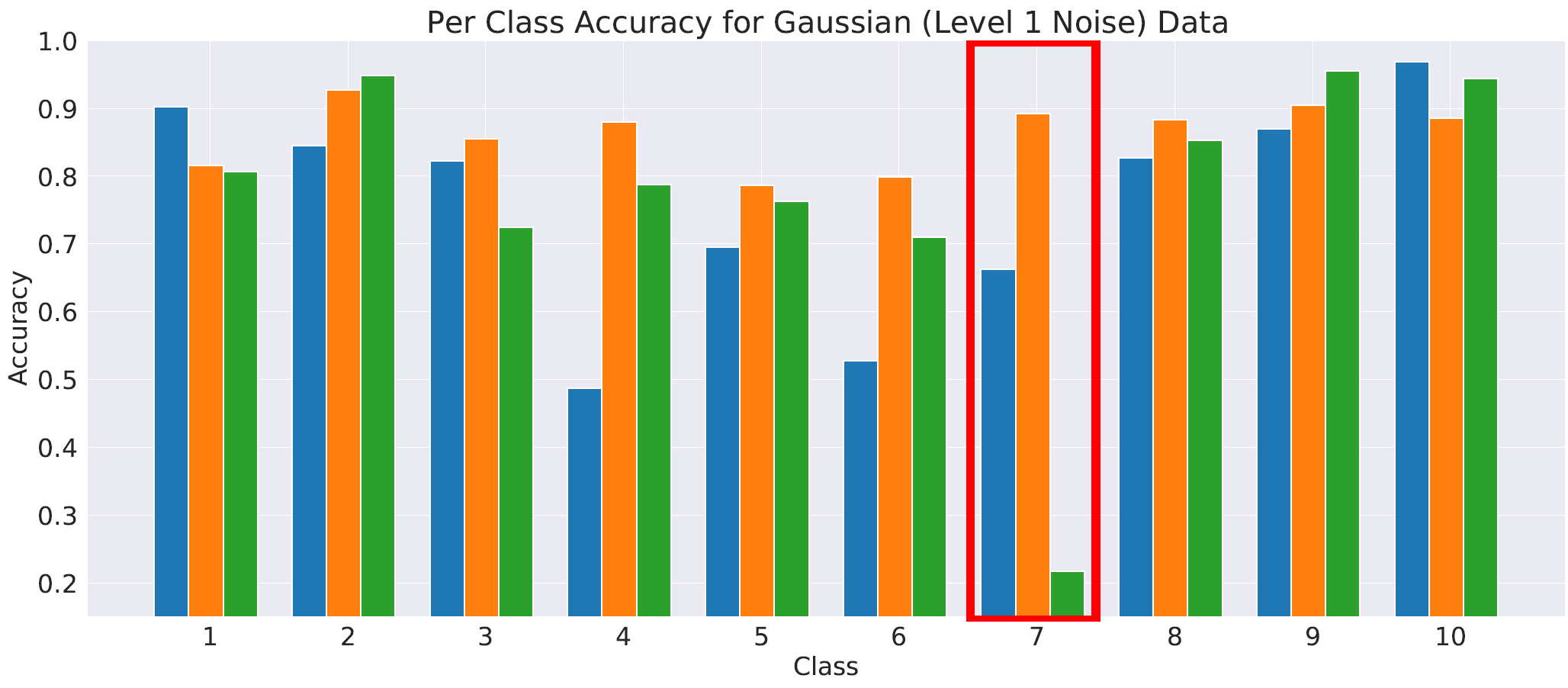}
    \includegraphics[width=0.25\textwidth]{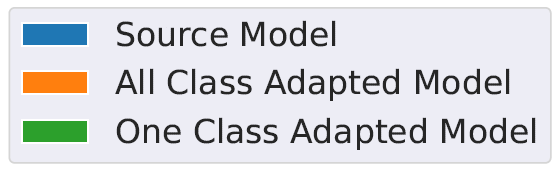}
\end{subfigure}
\caption{Effect of extreme class imbalance and covariate shift on test-time batch-normalized (TTN) performance: TTN mitigates distributional shift but greatly suffers from class imbalance. We show per-class accuracy plots for a Source model (blue), a class-balanced TTN model (orange), and a class-imbalanced (1-class) TTN model (green). Observe that the class-imbalanced TTN model performs very poorly on the most prevalent class label of the imbalanced adaptation set (label 7, red rectangle).}
\label{fig:CM_all}
\end{figure}

The main idea behind TTA in general, and TTN in particular, is that while label information is not available at test time, the unlabeled data can provide some information to estimate impact of domain shifts on neural network operations and internal representations. The typical setup is based on data being processed in batches, enabling assessment of distribution shifts between source and target domains. 

In order to implement TTA in these settings, TTN views a neural network 
$n$ as split into blocks separated by BatchNorm layers:
 \begin{equation}
      n = n_0\circ B_0^{s_0} \circ n_1 \circ B_1^{s_1}\cdots n_{K-1}\circ B_{K-1}^{s_{K-1}}\circ n_K,
 \end{equation}
where 
$n_0, \ldots, n_K$ are blocks (i.e., sub-networks) of hidden layers and $B_0, \ldots, B_{K-1}$ are batch normalization operators. Each BatchNorm layer modifies each neural activation by 
\begin{equation}
B(h(x),\mu_k,\sigma_k,\beta,\gamma) = \gamma \frac{h(x)-\mu_k}{\sigma_k} + \beta
\end{equation}
where $h(x)$ is the layer input batch, $\beta$ and $\gamma$ are parameters learned during the training process, and $\mu$ and $\sigma$ represent estimates of the mean and standard deviation of neuron activation over data. 

The main premise of the TTN approach is that changes in the distributions of activations of each neuron between source and target batches would predominantly be caused by unwanted covariate shifts, and therefore should be eliminated (TTN eliminates a diagonal affine shift exactly). However, this does not take into account other distribution shifts that \textit{should} affect, at the very least, the output distribution of the network. In particular, in many real-world applications, one cannot guarantee that classes will be well balanced in general and certainly at test time. Therefore it is often the case that the distribution of available labels during the training process, i.e., on the source domain, will differ from the one encountered at test time. 

Our proposed method to tackle this problem consists of three primary ingredients: 
\begin{itemize}
    \item Creating a channel-level score to indicate the sensitivity to individual classes
    \item Using a simple prior that linearly decays the number of selected channels in depth
    \item A prior on the label distribution, this can be uniform or derived from a pseudolabeling step
\end{itemize}

\begin{minipage}{0.45\textwidth}
  \begin{algorithm}[H]
    \SetAlFnt{\small}
    \DontPrintSemicolon
    \SetAlgoLined
    \LinesNumberedHidden
      \KwIn{Trained network $n$ with $K$ layers; source model statistics $\{\bm{\mu}_k^s,\bm{\sigma}_k^s\}_{k=1..K}$; 
      source domain data $D$}
        \For{$c$ in $C$ classes}{
            \textsc{Sample Data from Class $c$}\\
           $x_0\sim D_c$ \\
          \For{$k$ in $\{1,...,K\}$}{
            $\bm{x}_{k} = n_{k-1}(\bm{x}_{k-1})$ \\
             $\bm{\mu}_k^t,\bm{\sigma}_k^t$ $\leftarrow$ \textsc{Compute BN stats from~} $\bm{x}_{k}$\\
        
             \For{$f$ in $F$ channels}{
             \textsc{Scores}[c, k, f] = $W_2^2(\bm{\mu}_f^s,\bm{\sigma}_f^s,\bm{\mu}_f^t,\bm{\sigma}_f^t)$\\
             }}}
     \textbf{Output:}\textsc{Scores}
     \caption{Hybrid-TTN Post-Training}
    
  \end{algorithm}
\end{minipage}
   \hfill
\begin{minipage}{0.5\textwidth}
  \begin{algorithm}[H]
  \SetAlFnt{\small}
  \DontPrintSemicolon
       \SetAlgoLined
       \DontPrintSemicolon
       \LinesNumberedHidden
      \KwIn{Trained network $n$ with $K$ layers; source model statistics $\{\bm{\mu}_k^s,\bm{\sigma}_k^s\}_{k=1..K}$; target data batch $x_0$, label distribution prior $p_{1..c}$.}
       \For {$k$ in $\{1,...,K\}$}{
        $\bm{x}_{k} = f_{k-1}(\bm{x}_{k-1})$ \\
         $\bm{\mu}_k^t,\bm{\sigma}_k^t$ $\leftarrow$ \textsc{Compute BN stats from~} $\bm{x}_{k}$\\
         \For {$c$ in $C$ classes}{
         \For {$f$ in $F$ channels}{
             \textsc{Scores$_k$}[f] += $p_c*$\textsc{Scores}[c, k, f]
         }}
         $R= \frac{k}{K}$\\
         \textsc{topR-ind} $\leftarrow$ \textsc{Compute Top-R Index}(\textsc{Scores$_k$}) \\
         $\bm{m}_k[\textsc{topR-ind}] = 0$ \\
         $\bm{m}_k[\tilde{\textsc{topR-ind}}] = 1$ \\
         $\bm{\mu}_k^{hybrid} = \bm{m}_k\odot\bm{\mu}_k^t + (1-\bm{m}_k)\odot\bm{\mu}_k^s$ \\
         $\bm{\sigma}_k^{hybrid} = \bm{m}_k\odot\bm{\sigma}_k^t + (1-\bm{m}_k)\odot\bm{\sigma}_k^s$ \\
         $\bm{x}_k = B(\bm{x}_k,\bm{\mu}_k^{hybrid},\bm{\sigma}_k^{hybrid})$\\
     } 
     
     \textbf{Output:}$\{\bm{\mu}_k^{hybrid},\bm{\sigma}_k^{hybrid}\}_{k=1..K}$, $x_K$
     \caption{Hybrid-TTN Adaptation}
  \end{algorithm}
\end{minipage}

We now describe the three phases of the method:

\textbf{Storing Post-Training Statistics} To mitigate the adverse effects by TTN, our first step focuses on selecting channels for the adaptation process. We aim to identify channels that are primarily sensitive to covariate shifts while excluding those highly sensitive to shifts in the label distribution. This channel selection procedure operates using an initial phase that uses the labelled training set before the model is deployed to compute a set of scores
, completely independent of the target data
. For each class, denoted as $c$ out of $C$ total classes, we input a subset of data from the class, $D_c$, into a trained neural network with $K$ layers. Each layer $k$ within this network possesses source statistics $\mu_k^s$ and $\sigma_k^s$, and the target statistics for the single class, $\mu_k^t$ and $\sigma_k^t$, are computed. The Wasserstein distance is then calculated for the source and target distributions with $W_2^2(\{\mu^s,\sigma^s\},\{\mu^t,\sigma^t\}) = \|\mu^s-\mu^t\|^2+(\sigma^s - \sigma^t)^2$. This step is described in Algorithm 1.

 

     

\textbf{Test-Time Scoring and Masking} For each channel we now have a notion of the sensitivity to each class. We consider a prior $p(c)$ on the class distribution of the target data. For each channel we compute the weighted score of the Wasserstein distances obtained at end of training, with the weighting determined by the class distribution prior. Under a uniform prior on the target data we can simply sum the sensitivities for each class to get a score representing the sensitivity. 

At each layer we choose the top-$R$ channels for adaptation. Using the prior that later layers are more sensitive to label shifts, we decay $R$ with a simple linear function in depth. The algorithm is fully described in Algorithm 2. 

 

 In the method analyzed in our paper we focus on a simple method to obtain $p(c)$. 
Initially, Algorithm 2 is executed with a uniform prior to provide an initial corrected model. 
The network is prepared for adaptation using the uniform prior, the training data scores calculated in Algorithm 1, and the linear depth prior. By considering the uniform prior as weights to the individual class scores, selected channels in each layer are changed to use the target data batch statistics, with the remaining keeping the source model statistics. The model output is a first classification prediction, which can be further improved if we recalibrate the label distribution prior, using the first prediction.

\textbf{Re-scoring with Pseudolabels} 
Following the recomputation of the distribution prior using the first prediction, Algorithm 2 is rerun, incorporating the updated, non-uniform prior for further refinement, and giving the second round of predictions, which we consider our Hybrid-TTN output. This secondary application of Algorithm 2 can be seen as aiming to adjust the model prediction to exclude rare classes in the target distribution, unlike the first step which avoids updating channels sensitive to individual classes.

\section{Experiments and Results}
\label{sec:results}

In this section we discuss our experimental setup. We use three datasets in our evaluations: two popular benchmarks, CIFAR-10-C and ImageNet-1K-C, and a publicly available fatty liver ultrasound dataset.

\textit{CIFAR-10 and CIFAR-10-C.} We use the CIFAR-10 \citep{krizhevsky2009learning} dataset along with CIFAR-10-C \citep{hendrycks2019benchmarking}. CIFAR-10 is a small natural image dataset with 50k training images and 10k validation images. CIFAR-10-C contains corrupted versions of the CIFAR-10 Validation set at varying severities.
We train our models on the uncorrupted dataset.

\textit{ImageNet-1K and ImageNet-1K-C.} We use the ImageNet-1K \citep{imagenet_cvpr09} dataset along with ImageNet-1K-C \citep{hendrycks2019benchmarking}. ImageNet-1K is a large natural image dataset with 1.2 million training images and 50k validation images. ImageNet-1K-C, similarly to CIFAR-10-C, contains corrupted versions of the ImageNet-1K Validation set at varying severities. Both CIFAR-10-C and ImageNet-1K-C are popular as a measure of robustness to covariate shift. 

\textit{Liver Ultrasound.} We use, as the training data, a dataset of ultrasound abdominal images collected at a liver transplantation center from patients under suspicion of non-alcoholic fatty liver disease (NAFLD) \citep{vianna}. Images were collected between September 2011 to October 2020 with seven different devices, with varying settings, and by multiple operators. It contains 7,529 B-mode images from 57 control cases without NAFLD and 142 patients with NAFLD. We train our models for binary classification, i.e. detecting fatty liver versus non-fatty liver, on this dataset. For the test data in TTN, we use a publicly available dataset collected at a different institution \citep{byra}, with 550 images. The differences between the source and the target data are multiple. On the patients level, we have differences in population demographics and in the inclusion criteria, e.g., the 55 patients on the publicly available dataset are diagnosed with obesity, which impacts the wave propagation during ultrasound exams. On the examinations level, the second dataset was collected prospectively, with consistent settings and using a single ultrasound machine, which was not present in the source data collection. This creates a more homogeneous dataset and provides the covariate shift for the TTN method. Finally, the training dataset has 84\% of its images from cases of fatty liver, while the test dataset has 69\%, introducing the label distribution shift. For both datasets, all patients underwent liver biopsy and the steatosis grades in the pathology reports are used as the reference standard for diagnosis of fatty liver. In order to introduce higher levels of covariate shift akin to CIFAR-10 and ImageNet-1K evaluations, we added speckle noise to the target dataset, with varying levels of corruption.

\textbf{Architecture Details}
On CIFAR-10 we train a Resnet-26 model as defined in \citep{he2016deep}. For ImageNet-1K we use a pre-trained Resnet-18 model. For the Liver Ultrasound data, we train a Resnet-18 model. 
Training details are described on Appendix \ref{sec:app_train}.

\noindent\textbf{Adaptation Details} We focus on the source-free domain adaptation setting where adaptation is done offline on a single batch without affecting the deployed model \citep{liang2023ttasurvey}, avoiding issues with catastrophic forgetting \citep{eata}. For CIFAR-10-C and ImageNet-1K-C we use a batch size of 500 for the experiments (sampled over multiple seeds). For Liver Ultrasound, the batch size is the length of the target set. For the Hybrid-TTN, the percentage of not-adapted channels by layer is defined by a linear function starting at 0\% at the first BatchNorm layer and ending at 100\% at the last layer. The step of this linear function changes by model (e.g., Resnet-26 has more BatchNorm layers than Resnet-18), but it does not introduce any hyperparameters. Our method is hyperparameter free and appplied on all datasets and settings in the same way. 
More adaptation details are described in the Appendix \ref{sec:app_adap}.

\noindent\textbf{Target Label Distributions} In order to analyze label distribution shifts we consider two approaches for constructing class imbalanced target distributions, while we keep the source training balanced. 
The first approach is 
to randomly select $N$ classes from the set of all classes and constructing a batch with the specified $N$ classes. We also use a more natural imbalanced dataset simulation construction technique popular in the federated learning literature which samples from a Dirichlet distribution with parameter $\alpha$ \citep{li2021federated}. Smaller $\alpha$ will yield more sample batches with higher imbalances. This allows us to mimic a more natural scenario where the label distribution is not only varying in classes but proportions. In Liver Ultrasound, we use the original target distribution, which is different from the original source distribution, alongside a balanced distribution, and severely imbalanced cases.

\subsection{Label Distribution Shift Impact on TTN}

We first illustrate the potential pitfalls of TTN on several examples of label distribution shift. Using the CIFAR-10 dataset we show the effect on TTN of an extreme version of label distribution shift in Figure~\ref{fig:CM_all}. Specifically, we adapt on batches containing one class or all classes (as can be seen in the label distributions shown in Figure~\ref{fig:CM_all}). We observe in the adaptation of one class batches that the accuracy of TTN degrades severely compared to the source model in cases where there is a covariate shift (application of Gaussian noise to the input) but also \textit{in the absence of covariate shift}. The latter observation is also confirmed for ultrasound data where we modify the target label proportion of the test data coming from the same distribution, but without adding noise (in Appendix 
\ref{sec:app_us}). 
This catastrophic failure, even without covariate shift, is clearly undesired when operating in a realistic deployment environment where the test set label distribution is unknown and can vary for each incoming batch.

\begin{figure}[h]
\centering
    \includegraphics[width=0.45\linewidth]{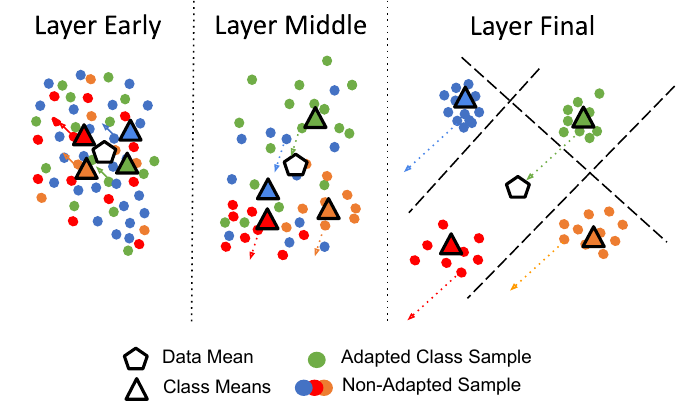}
    \caption{Illustration of explanation for depth-wise behaviour under label distribution shift. We consider one class mean (green) which is shifted towards the data mean, as would be the case in a highly imbalanced setting. Classes are not well separated in early layers and thus shifts in any mean are relatively small and non-intrusive. In later layers classes are well separated and a large shift of points from one mean towards the data mean is likely to cross a decision boundary. Data points in other classes moving away from the data mean are less likely to cross a decision boundary.}
    \label{fig:DiagramTTN}
\end{figure}

Furthermore, we can observe from the per-class performance evaluations of the adapted models that the majority class is disproportionately affected by TTN, and that classes underrepresented (or not present) will, unintuitively, experience less degradation or even slightly improved performance compared to source model. We can obtain intuition about this latter behavior by considering a strong classification model where representations from each class will be well separated and clustered in the final representation layer (see Figure~\ref{fig:DiagramTTN}). Note that this is an assumption that is correct for most modern neural network architectures, as it is a consequence of the neural collapse phenomenon first described in \citet{PapyanDonoho2020}. A biased BatchNorm computation will compute the mean of a specific label cluster and re-center the data by this mean. Now, consider a neural network layer whose activations are tightly clustered, such that each cluster corresponds to one label. Further, assume that the different cluster centers occupy approximately orthogonal directions in this space. As mentioned above, such quasi-orthogonality is not an uncommon property in high dimensions \citep{PapyanDonoho2020}. In this setting, linear decision directions would align with cluster means. Subtracting a cluster mean instead of a class-balanced batch mean would then significantly move the label cluster along its decision direction and will likely lead to some examples passing below the classification threshold. Other classes will be barely affected due to the orthogonality of the cluster centers and hence of the decision boundaries.

\subsection{Label Sensitive Channel Level Adaptation}
As mentioned previously, many works in deep learning have shown that supervised trained neural networks become more specialized as depth of a network increases \citep{zeiler2014visualizing, belilovsky2019greedy, oyallon2017building} and that classes progressively become more separated with depth. This suggests the later layers are more likely to have channels which are very sensitive to specific classes. 

\begin{figure}[h]
\centering
\includegraphics[width=0.4\textwidth]{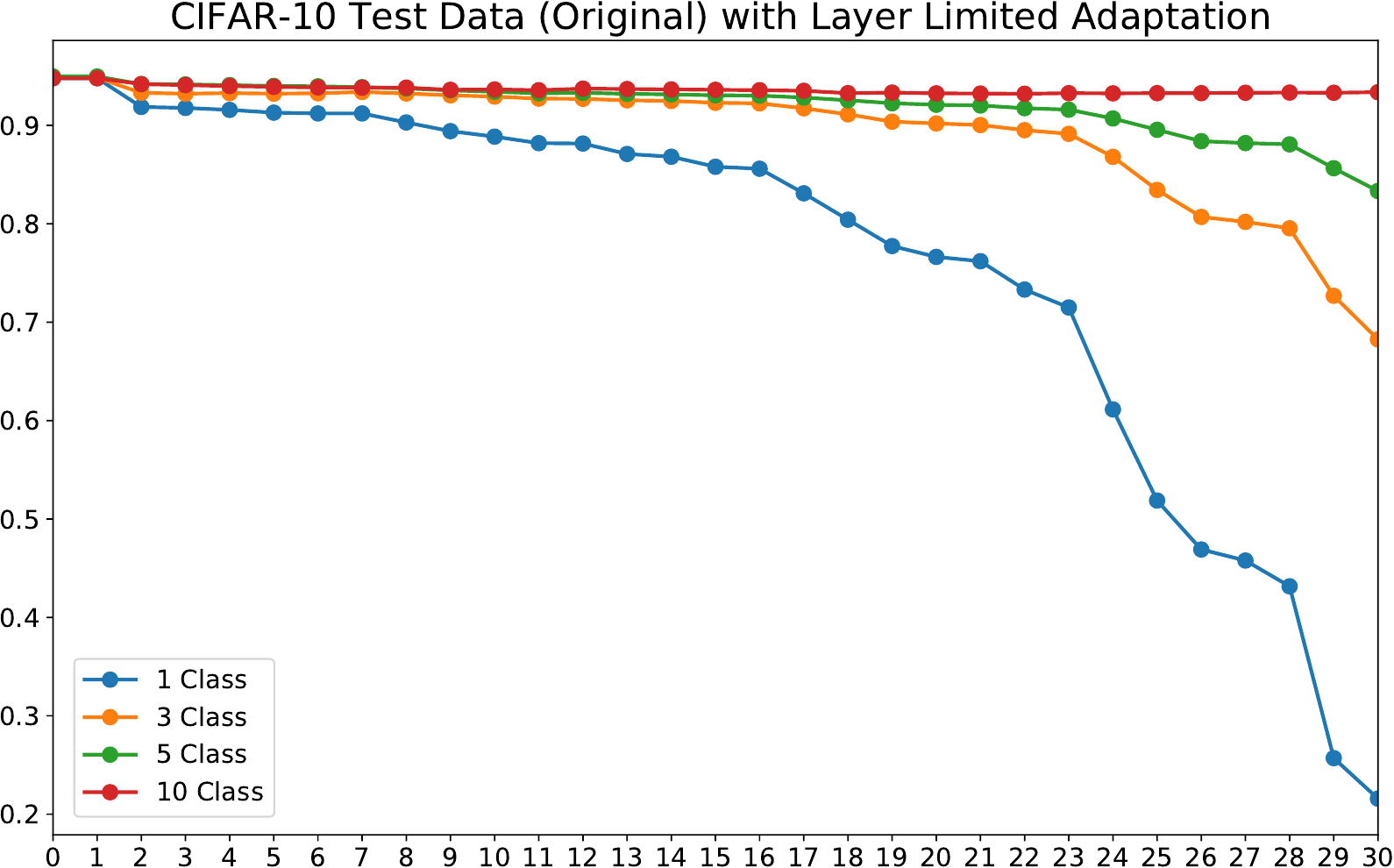}
\includegraphics[width =0.4\textwidth]{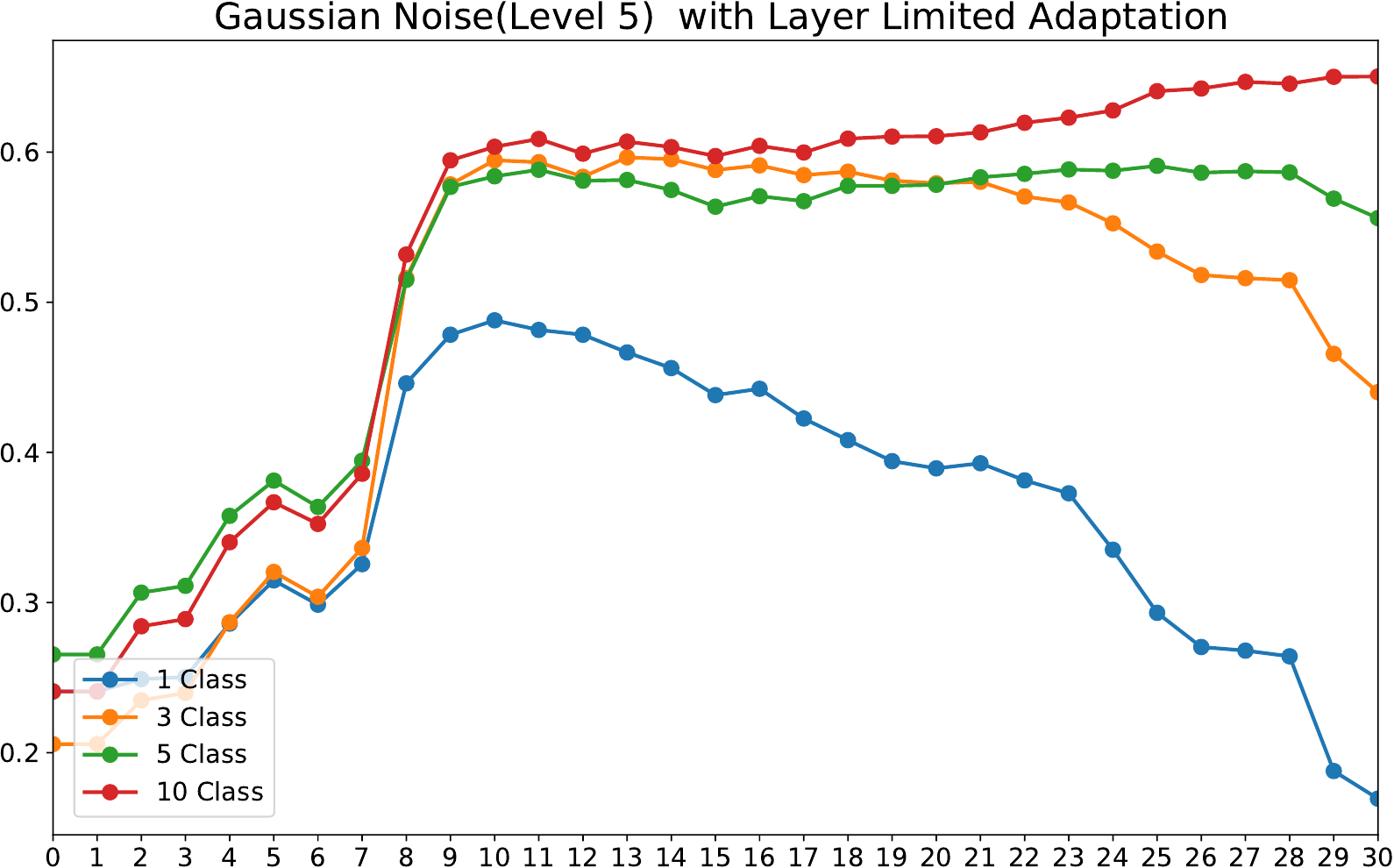}
\caption{On CIFAR-10 we adapt models only up to the layer shown on the x-axis, the y-axis showing the accuracy on the target data. We consider label distributions with all (10) classes as well as 5,3, and 1 randomly selected and balanced classes. 
We observe that adapting later layers has an outsized role in the catastrophic collapse due to TTN. }
\label{fig:layerlimitedcifar}
\end{figure}

\textbf{Adaptation of Early Layers}
To further confirm that depth can have an effect on the behavior of TTN and taking into account that earlier layers tend to be less specialized, we now consider only adapting earlier parts of the network and study this effect on the classification performance under label distribution shift. We perform experiments on layer-limited adaptation for a Resnet-26 model on CIFAR-10 both with and without noise. Our results are shown in in Figure~\ref{fig:layerlimitedcifar}. Specifically, \textit{for each index on the x-axis we perform test-time normalization of the model up to this layer}. Consequently, the extremes of the x-axis (0 and, here, 30) correspond to Source and TTN models respectively. First, we observe that on the non-corrupted train data the performance of class-imbalanced data degrades gradually at first and increasingly faster towards the later layers. This suggests that later layers can cause a large degradation. Secondly, for corrupted data we observe that adapting up to earlier layers can allow enough label distribution invariance to provide benefits under covariate shift. For example, models that are adapted up to layers 8 through 15 can perform much better than source models under covariate shift but without label distribution shift, while also being able to perform much better than fully adapted and often source models in various label distribution shift scenarios. This suggests limiting adaptation in later layers can be beneficial. On the other hand, it is not obvious a priori which layer should be adapted. Furthermore, different classes can have more fine-grained effects, for example, some classes may lead to over-sensitivity in certain neurons or channels in a layer than others. 

\begin{figure}[h]
    \centering
    \includegraphics[width=0.35\linewidth]{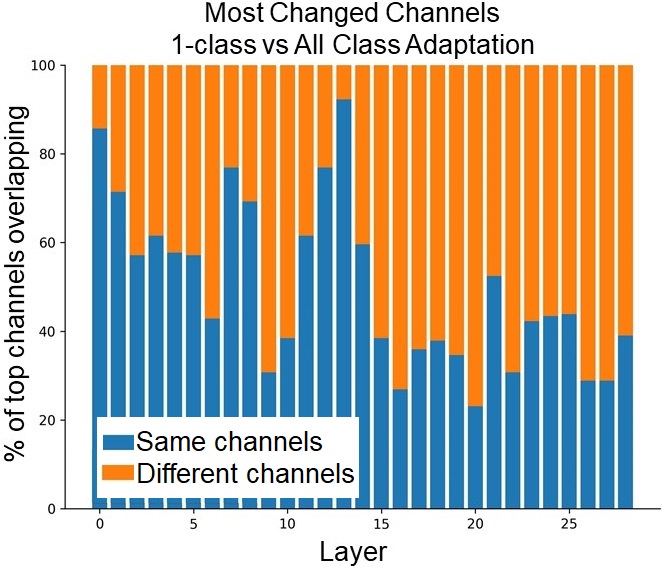}
    \includegraphics[width=0.35\linewidth]{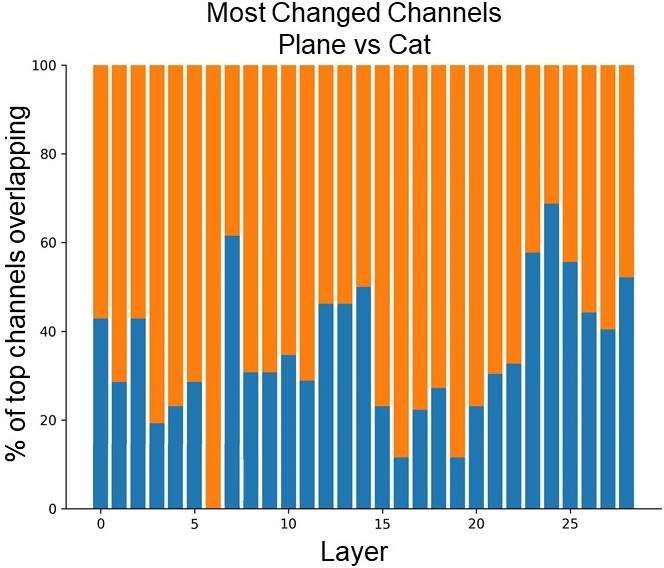}
\caption{We compute the BN statistics for all-class and multiple 1-class cases and compare agreement on channels which are most adapted (as measured by top-10\% Wasserstein distance). The percentage of overlapping channels that presents the highest Wasserstein distance (between source and adapted models) in both tasks is represented by the blue bars, with orange bars representing the percentage of channels among those with the highest Wassertein distance that are not the same between the two evaluated cases.} 
\label{fig:topK}
\end{figure}

\noindent\textbf{Adaptive Channel Selection}
We now study how different class distributions affect the channels that are most adapted. We use a similar setup as in the previous example and consider within a layer the Wasserstein distance as a metric for ranking the channels which would be most changed by TTN adaptation. Subsequently, we take the channels in the top 10\% for each layer and compare them for the cases of 1-class vs all-class adaptation, and 1-class vs another. Specifically, we determine how many channels in the top adapted channels for each batch (single classes, all classes) overlap. The results are given in Figure ~\ref{fig:topK}. 
We see from the first diagram that the channels that change the most when adapting to just 1-class versus all-class tend to have more agreement in the earlier layers, further supporting that later layers have many more channels sensitive to label distribution shifts.

However, we can see if we compare two different classes (e.g., Plane vs Cat) that the adaptation can vary substantially, throughout the network and even towards the last layers. This suggest that depending on the class distribution, different channels may be more sensitive to different classes. This supports our dynamic strategy that selects the most sensitive channels to remove from the adaptation set.
\subsection{Evaluating Hybrid-TTN}
\begin{figure*}[h]
    \centering
    \includegraphics[width=1\linewidth]{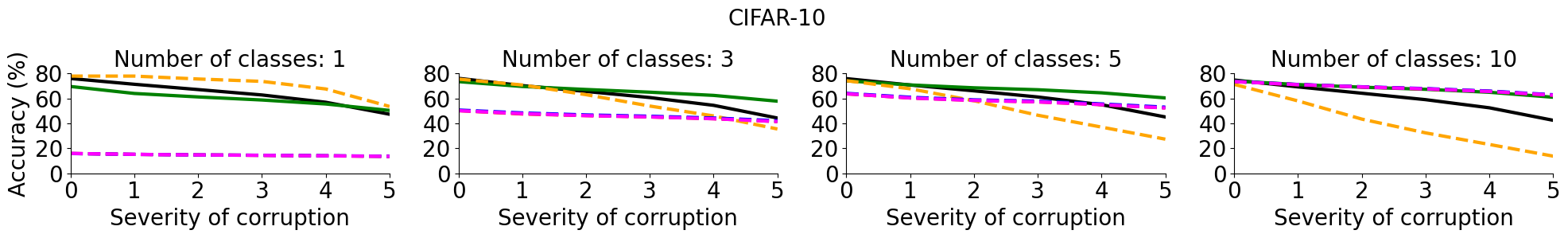}
    \includegraphics[width=1\linewidth]{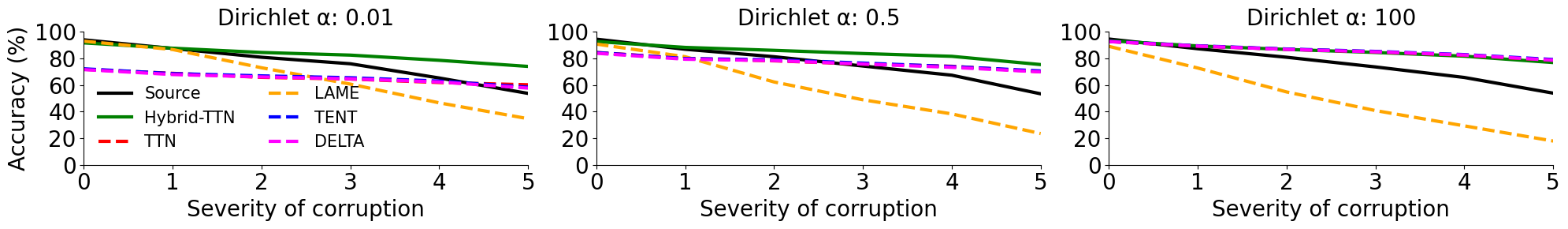}  
\caption{CIFAR-10 evaluations on multiple label shifted distributions and covariate shifts (all corruptions) with different degrees of label imbalance. We show the source model accuracy and the improvement (or degradation) with our proposed method and baseline. We observe that the proposed method provides benefits when there is no covariate shift, while avoiding catastrophic failures and allowing benefits over source when there are label distribution shifts.} 
\label{fig:cifar_results}
\end{figure*}

We now use Hybrid-TTN method on a variety of target datasets, covariate shifts, and label distribution shifts. We demonstrate that Hybrid-TTN can provide a good trade-off in being able to adapt to covariate shift without experiencing catastrophic failure due to label distribution shift. Our results are shown in Figures \ref{fig:cifar_results}, \ref{fig:imagenet_results}, and \ref{fig:ultrasound_results}, and 
more 
detailed tables are available in the Appendix 
\ref{sec:app_results} and \ref{sec:app_corruptions}. Here we demonstrate various degrees of covariate and label distribution shift, and the gain or loss as compared to the source model performance.

\begin{figure*}[h]
    \centering
    \includegraphics[width=0.99\linewidth]{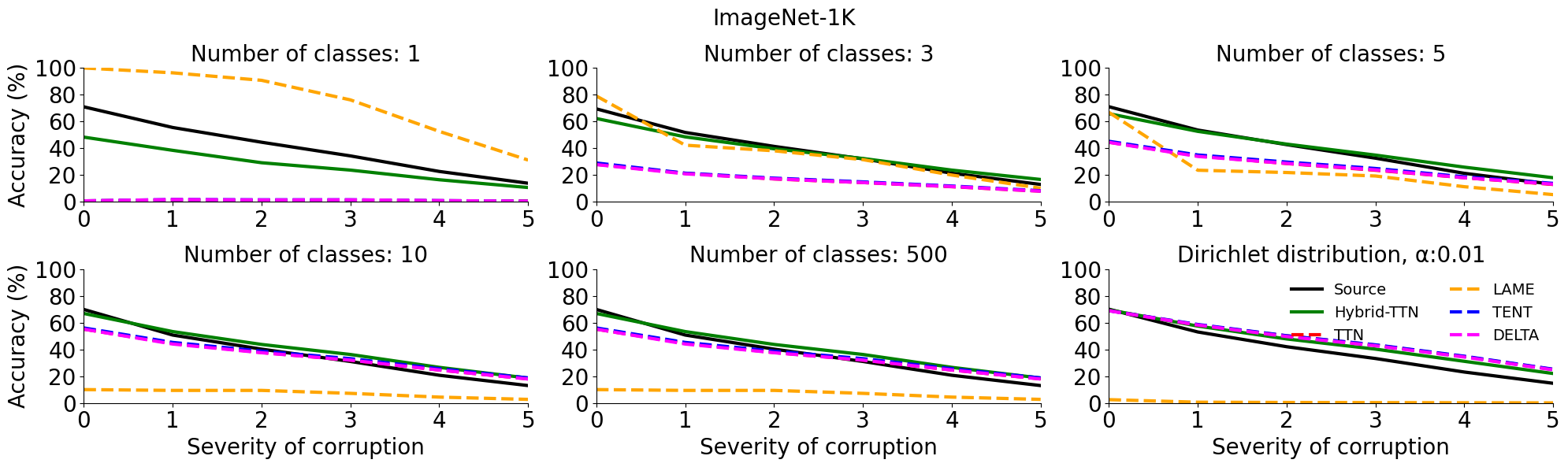}
\caption{ImageNet-1K evaluations on multiple label shifted distributions and covariate shifts (all corruptions) with different degrees of label imbalance. 
We observe that the proposed method provides benefits when there is covariate shift, while avoiding catastrophic failures when there are label distribution shifts.} 
\label{fig:imagenet_results}
\end{figure*}

For CIFAR-10 and ImageNet-1K we observe that Hybrid-TTN is able to control the degradation in performance under a variety of label distribution shifts in the absence of covariate shift, while TTN, TENT, LAME, and Delta can present severe decreases in performance depending on the label distribution. When corruptions are applied, Hybrid-TTN is able to bring benefits (substantially improve on source model) for the original distribution. Unlike other methods, it is able to handle the label distribution shift, in many cases avoiding catastrophic failure, and in a variety of combinations of severe label and covariate shift improving over the source model. We observe that LAME, which is designed for the label distribution shift problem, is able to perform well in the severes case of batches with a single class, but performs poorly in the cases that the data is balanced. 

We also evaluate on a realistic ultrasound dataset, in Figure \ref{fig:ultrasound_results}. Under a shift to a target dataset from a completely different site, we first observe that most methods can maintain or improve the performance compared to the source model in both original distribution and the balanced scenarios, the exception being LAME. This can be compared to the balanced distributions of the results for CIFAR-10-C, where the same methods achieved good performance when dealing with several levels of corruption. 

\begin{figure}
    \centering
    \includegraphics[width=1\linewidth]{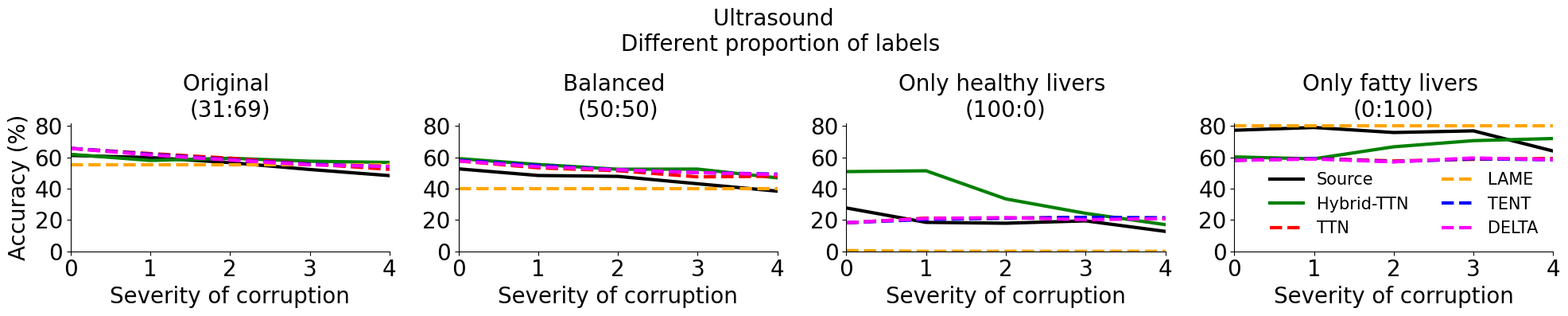}
\caption{Evaluation on ultrasound data. Binary classification, subjects with and without fatty liver disease. Models were trained on one site and are evaluated on a publicly available dataset \citep{byra}. We show evaluations on targets with shifts in label distribution. We observe that the proposed method is able to control the degradation in performance under severe label distribution shift.}
    \label{fig:ultrasound_results}
\end{figure}

When we consider more severe imbalance scenarios, such as only using patients without fatty liver and only patients with fatty liver, we see varying results. Hybrid-TTN consistently outperforms the source model in the case of healthy livers, but in the fatty livers case it can only surpass the source model when severe corruption is present. LAME is consistently near 100\% accuracy in the fatty liver scenario and 0\% accuracy in the healthy liver scenario, indicating that the method might be biased towards the training distribution. The other methods are stable across levels of corruption but have diminishing returns.

\noindent\textbf{Trade-offs of TTA} We emphasize that the goal of Hybrid-TTN is to avoid severe failures allowing a user to deploy TTA without concern that it might be counterproductive compared to using just the source model. To illustrate this we also rank the evaluated methods within each dataset for all scenarios (label shift, covariate shift) and compute the median rank. The results are presented in Table~\ref{tab:performance}. We first observe that when viewed under this evaluation, the source model for CIFAR-10 and ImageNet-1K outperforms all the other TTA methods, indeed suggesting that TTA must be applied with extreme caution. To illustrate this point, we can observe that LAME can result in high accuracy increase in specific scenarios, but not on most cases. Similarly, other techniques are more suited to specific use-cases. On the other hand, Hybrid-TTN \textit{is the only TTA method able to obtain a substantially better median rank compared to the source model}, suggesting it is conservative while providing benefits under covariate shift making it a useful method for real-world deployment.  

\begin{table}[h]
\centering
\footnotesize
\begin{tabular}{|c|c|c|c|}
\hline
& \multicolumn{3}{c|}{Ranking ($\downarrow$)} \\ \cline{2-4} 
Technique & CIFAR-10 & ImageNet-1K & Ultrasound \\ \hline
Source & 2.0 & 2.8 & 5.0 \\
TTN & 4.5 & 4.4 & 3.5 \\ 
TENT & 3.0 & 2.8 & 3.3 \\ 
LAME & 6.0 & 6.0 & 6.0 \\ 
DELTA & 4.5 & 4.3 & 3.0 \\
\textbf{Hybrid-TTN} & \textbf{1.0} & \textbf{2.3} & \textbf{1.5} \\ \hline
\end{tabular}
\caption{Median ranking for each technique across the three different datasets. Lower is better. Across multiple distributions and degrees of corruption, our proposed method offers the best overall performance in multiple tasks and is the only TTA method to show consistent utility from adaptation.}
\label{tab:performance}
\end{table}

\textbf{Computational Complexity} In terms of computational complexity, we note that the scores are computed in Algorithm 1 prior to deployment, thus Algorithm 1 is run only once while Algorithm 2 can be used for any new data. Algorithm 2 adds minimal overhead to a standard forward pass. We benchmark our implementation of the Hybrid-TTN approach and note it takes $0.89 \pm 0.01$ seconds to output its predictions, compared to $0.39 \pm 0.003$ seconds for TTN which is the lowest overhead TTA method. This could be explained by the two passes of Algorithm 2, using a different class prior each time. A comparison of runtimes across different methods evaluated is presented in the Appendix \ref{sec:app_runtimes}.

\subsection{Ablation of Channel Selection}
We perform an ablation study aimed at investigating the efficacy of the different elements of the channel selection strategy within Hybrid-TTN. 
The ablation experiments for a single class across all corruptions is presented in Table \ref{tab:cifar_ablation}.

\textbf{Depth Prior Only} We first consider the importance of the scoring function compared to the linear decaying thresholding. We denote this ablation the HybridTTN+Random Channel Scores. Specifically we replace the scores used for thresholding in Algorithm 2 (line 10) with random scores. This decouples the scoring obtained from Algorithm 1, with the prior on depth wise adaptation. We observe that we obtain substantial improvement over just standard TTN but that the performance degrades severely compared to our HybridTTN.

\textbf{Class Prior} We now consider the importance of the pseudo-labeling step used to obtain a class prior from the target data  described in Section \ref{sec:method}. Specifically, we evaluate the performance of just using a uniform prior, we observe this improves the performance from the random channel scores, but substantially underperforms Hybrid-TTN. Finally we also show the performance of an oracle class prior, where the exact target data label distribution is given. We note that for none or low corruption cases this gives performance comparable to Hybrid-TTN, further affirming our approach. As the corruption becomes more severe the performance gap between the oracle and our method increases. 
This effect was also observed in additional experiments considering different CNN architectures and CIFAR-100, detailed in the Appendix \ref{sec:more_ablation}.

\begin{table*}[h]
\footnotesize
    \centering
    \begin{tabular}{|c|c|c|c|c|c|c|}\hline
     & \multicolumn{6}{|c|}{Covariate Shift} \\\hline
     & \multicolumn{6}{|c|}{Accuracy (\% or $\Delta$\%)  } \\\hline
    & Original & Corruption-1 & Corruption-2 & Corruption-3 & Corruption-4 & Corruption-5\\\hline
     Source & 95.0 ± 2.4 & 89.2 ± 9.9 & 83.9 ± 12.5 & 78.5 ± 16.5 & 71.2 ± 20.5 & 59.2 ± 26.5\\\cdashline{2-7}[2pt/2pt]
     TTN & -75.1 ± 3.9 & -70.2 ± 9.7 & -65.5 ± 12.0 & -60.5 ± 16.0 & -53.7 ± 19.7 & -42.2 ± 25.9\\
  Hybrid-TTN & -8.0 ± 3.1 & -9.2 ± 5.7 & -7.4 ± 7.9 & -5.1 ± 10.8 & -1.7 ± 14.9 & +3.8 ± 22.1\\\cdashline{2-7}[2pt/2pt]
  + Random Channel Scores & -21.9 ± 6.9 & -25.7 ± 11.6 & -22.6 ± 14.3 & -18.4 ± 18.3 & -13.2 ± 22.6 & -5.2 ± 28.1 \\
   + Uniform Prior Class Prior & -14.1 ± 4.5 & -15.4 ± 6.6 & -13.8 ± 8.5 & -11.1 ± 11.8 & -7.4 ± 15.5 & -1.3 ± 22.6 \\
   + Oracle Class Prior & -7.8 ± 3.2 & -7.5 ± 5.6 & -5.5 ± 7.6 & -2.8 ± 11.0 & +1.6 ± 15.2 & +8.1 ± 22.1 \\
   
  \hline
   
    \end{tabular}
        \caption{Results for the ablation study. Each technique delta accuracy is in comparison to the source results. We use CIFAR-10 with batches from a \textit{single class} and different degrees of covariate shift. We show that our Wassertstein scoring and class prior estimates are both beneficial to Hybrid-TTN.}
        \label{tab:cifar_ablation}

\end{table*}

\vspace{-15pt}
\section{Limitations} 
A limitation of our proposed technique is that it is designed for architectures featuring batch normalization layers. However, it is noteworthy that this is still a popular kind of architecture in the TTA literature \citep{delta}. Furthermore, future work can adapt our channel selection criteria to other methods, by using a different kind of scoring function and avoiding adaptation (e.g. to entropy minimization) of selected channels. Another limitation is that we assume access to the post-training but pre-deployment access to the source data, however note that this is practical in most cases except where the entity training the model does not know of the eventual application of TTA to the model.  

\vspace{-5pt}
\section{Conclusions}
\label{sec:discussion}

We have studied a popular batch-level Test-time Adaptation method in the context of label distribution shift. We observed that in realistic scenarios where batches at deployment time have label distribution shifts, this method can fail catastrophically. A comparison in the offline setting indicates that common test-time adaptation techniques are detrimental to classification, contingent upon shifts in label distribution.
We proposed a direction for solving this problem to keep the benefits of adaptation without risking catastrophic failure due to label shift. 
Compared to similar works, our proposed technique was designed to be generic and easily applied without hyperparameter tuning, as we assume a fixed linear depth prior and a uniform distribution prior which are applied to all datasets and architectures studied. 
Moreover, our results indicate that the challenge of domain adaptation in the biomedical field, specifically in diagnosis of fatty liver using grayscale ultrasound images, can be addressed by the proposed Hybrid-TTN method, reducing risks in the deployment of a trained framework in different populations.




\subsubsection*{Acknowledgments} \vspace{-8pt}
This work was supported by grants from the Institute of Data Valorization (IVADO PRF3) to A.T., G.W. and E.B. 
A.T. acknowledges support from the Fonds de Recherche du Québec – Santé (FRQS), Fonds de Recherche du Québec  - Nature et technologies (FRQNT)
 and the Fondation de l’Association des Radiologistes du Québec (FARQ) Clinical Research Scholarship–Senior Salary Award (FRQS-ARQ no. 298509). 
G.W. acknowledges support from the Canada CIFAR AI Chair. 
E.B. and G.W. acknowledge funding from Fonds de recherche du Québec — Nature et technologies - NOVA (2023-NOVA-329759 and 2023-NOVA-329125).



\newpage 

\bibliography{collas2024_conference}
\bibliographystyle{collas2024_conference}
\newpage 

\appendix
\section{Appendix}
This appendix includes:

\begin{itemize}
\item Training details (Sec. \ref{sec:app_train})
\item Adaptation details (Sec. \ref{sec:app_adap}) 
\item Ultrasound failure case without domain shift (Sec. \ref{sec:app_us})
\item Runtime comparison across techniques (Sec. \ref{sec:app_runtimes})
\item Comparison to different architectures and CIFAR-100 (Sec. \ref{sec:more_ablation})
\item Detailed tables for CIFAR-10-C, ImageNet-1K-C and Ultrasound results (Sec. \ref{sec:app_results})
\item CIFAR-10 results by corruption type (Sec. \ref{sec:app_corruptions})
\end{itemize}

\subsection{Training details} \label{sec:app_train}
On CIFAR-10 we train a Resnet-26 model as defined in \cite{he2016deep}. We use an
SGD optimizer with a batch size of 128. An initial learning rate set to 0.1 is used in combination with a cosine annealing schedule \citep{DBLP:journals/corr/LoshchilovH16a} trained over 200 epochs. Weight decay set to 5e-4 is used along with momentum set to 0.9 \citep{DBLP:journals/corr/ZagoruykoK16}. Standard augmentation uses random crop of size 32 with 4 padding, and random horizontal flips. 

For the Liver Ultrasound data, we train a Resnet-18 model with similar configurations, using a SGD optimizer with learning rate set to 0.001 and a batch size of 32. Differences in the models configurations can be explained by the computational constraints of the Hospital Research Center where the images are secured.


\subsection{Adaptation details} \label{sec:app_adap}
For TTN, we use the original configuration, resetting batch normalization layers (BatchNorm) running stats. The difference between TTN and Hybrid-TTN is that for the former we adapt all channels in all BatchNorm during test-time, without our proposed channel selection.

For TENT, original configuration was followed. In addition, the adaptation for each batch was repeated for 10 iterations.
In implementing LAME, we adhered to the default configuration provided in the officially released code for achieving best model performance. As per the specified configuration, we chose the radial basis function kernel for the affinity matrix with K-nearest neighbors and $\sigma$ set to 5 and 1 respectively.
We also used the same hyperparameters for optimizer as specified in the mentioned configuration.

To maintain consistency in our comparative analysis, we executed the experiments we adapted DELTA to our setup in which we had one test batch of a specific size. Notably, DELTA is tailored for online setups, excelling in processing multiple batches within a test stream. 


\subsection{Example of failure case in Ultrasound} \label{sec:app_us}
In Table \ref{tab:ultrasound1} we note a significant decline in TTN accuracy compared to the source model, not only in instances with covariate shifts but also without any covariate shift. This phenomenon can be seen in the ultrasound data, where we adjust the proportion of target labels in the test data sourced from the same distribution as the source data. 
\begin{table}[h]
\small
    \centering
    \begin{tabular}{c|c|c}
&	\multicolumn{2}{c}{Accuracy (\%)} \\\hline
Labels proportion	&	Source	&	TTN	\\\hline

Original (16\%/84\%)	&	82.3 	& 83.1\\
0\%/100\% 	&	92.2 	&	84.4 \\
50\%/50\% 	&	71.9 	&	67.2 \\
100\%/0\% 	&	51.6 	&	21.9
\end{tabular}
      \caption{Illustration of failure case on binary classification of medical ultrasound data (detecting presence of fatty liver). If TTN is deployed on target data from the same domain (same hospital, demographics) with the original training proportions (16\%/84\%) performance is maintained or slightly improved. On the other hand, if the target batch presented has a label distribution shift we see substantial degradation.}
    \label{tab:ultrasound1}
\end{table}

\subsection{Adaptation runtimes} \label{sec:app_runtimes}
Below we present the table for the runtime of each investigated test-time adaptation technique. Values are averaged over 50 random seeds and each experiment is for the process time of a batch of size 500 of CIFAR-10 images.

\begin{table}[h]
    \small
    \centering
    \begin{tabular}{c|c}
    \hline
    \textbf{Algorithm} & \textbf{Runtime (seconds)} \\ \hline
    Source & 0.23 $\pm$ 0.002 \\ 
    TTN & 0.39 $\pm$ 0.003 \\ 
    TENT & 3.65 $\pm$ 0.01 \\ 
    LAME & 0.83 $\pm$ 0.03 \\ 
    Delta & 0.92 $\pm$ 0.04 \\ 
    Hybrid-TTN (Uniform prior) & 0.49 $\pm$ 0.01 \\
    Hybrid-TTN & 0.89 $\pm$ 0.01 \\ \hline
    \end{tabular}
        \caption{Runtime of algorithms, in seconds.}
        \label{tab:runtime}
\end{table}

\subsection{Additional ablation experiments} \label{sec:more_ablation}
In this section, we show ablation results for using Resnet-50 and Resnext-29 with the same configurations as used for the original Resnet-26 in CIFAR-10. Moreover, we run the three architectures using CIFAR-100 as source and CIFAR-100-C as the target domain, instead of CIFAR-10 as used in the paper. The results showed below are only for the extremely imbalanced scenario, with only one class on the target set, and for extremely corrupted cases, with severity level 5 across all corruptions.

\begin{table}[h]
    \centering
    \begin{tabular}{c|c|c|c}
    \hline
    \multicolumn{4}{c}{\textbf{CIFAR-10 to CIFAR-10-C}} \\ \hline
    \textbf{Method} & \textbf{Resnet50} & \textbf{Resnet26} & \textbf{Resnext29} \\ \hline
    Source & 60.1 $\pm$ 26.4 & 59.2 $\pm$ 26.5 & 53.1 $\pm$ 30.1 \\ \cdashline{1-4}[2pt/2pt]
    TTN & -42.7 $\pm$ 25.8 & -42.2 $\pm$ 25.9 & -37.9 $\pm$ 29.7 \\     Random channel selection  & -20.1 $\pm$ 26.6 & -14.1 $\pm$ 27.6 & +3.4 $\pm$ 25.2 \\     
    Hybrid-TTN (Uniform prior) & -2.1 $\pm$ 24.9 & -1.3 $\pm$ 22.6 & +7.7 $\pm$ 19.0 \\     
    Hybrid-TTN (Proposed) & +1.9 $\pm$ 24.3 & +3.8 $\pm$ 22.1 & +8.5 $\pm$ 19.0 \\    
    Hybrid-TTN (Oracle prior) & +5.2 $\pm$ 24.4 & +8.1 $\pm$ 22.2 & +16.5 $\pm$ 20.7 \\ \hline
    \multicolumn{4}{c}{\textbf{CIFAR-100 to CIFAR-100-C}} \\ \hline
    \textbf{Method} & \textbf{Resnet50} & \textbf{Resnet26} & \textbf{Resnext29} \\ \hline
    Source & 26.1 $\pm$ 25.2 & 24.0 $\pm$ 25.0 & 25.7 $\pm$ 25.4 \\ \cdashline{1-4}[2pt/2pt]
    TTN & -23.2 $\pm$ 25.2 & -21.5 $\pm$ 25.1 & -22.5 $\pm$ 25.3 \\ 
    Random channel selection & -7.9 $\pm$ 23.5 & -8.4 $\pm$ 22.3 & -2.8 $\pm$ 21.8 \\ 
    Hybrid-TTN (Uniform prior) & +1.0 $\pm$ 19.3 & -0.8 $\pm$ 19.9 & +2.2 $\pm$ 15.8 \\ 
    Hybrid-TTN (Proposed) & +6.1 $\pm$ 18.1 & +3.4 $\pm$ 19.2 & +4.5 $\pm$ 15.4 \\ 
    Hybrid-TTN (Oracle prior) & +13.9 $\pm$ 19.9 & +13.7 $\pm$ 21.0 & +13.0 $\pm$ 16.5 \\ \hline
    \end{tabular}
        \caption{Ablation study using different models on CIFAR-10 and CIFAR-100. Values for Source model are accuracy, in percentages. Values for different ablated methods are the difference in accuracy between each method and the Source model. Uniform and Oracle priors are considering balanced target sets and true label distribution, respectively. Proposed method uses the pseudo-labels obtained with Uniform prior.}
    \label{tab:ablation_performance}
\end{table}

\subsection{Detailed results} \label{sec:app_results}
Below, we present the tables for all of our experiments 
for CIFAR-10 (Table \ref{tab:cifar_main}), Imagenet-1K (Table \ref{tab:imagenet}), and Ultrasound (Table \ref{tab:ultrasound_deltas}). We report the accuracy ($\%$) for the source model, and changes in accuracy ($\Delta \%$) for each different method.

\begin{table*}[]
\small
    \centering
    \begin{tabular}{|c|c|c|c|c|c|c|c|}\hline
    \multicolumn{2}{|c}{} & \multicolumn{6}{|c|}{Covariate Shift} \\\hline
    \multicolumn{2}{|c}{} & \multicolumn{6}{|c|}{Accuracy (\% or $\Delta$\%)  } \\\hline
    \multicolumn{2}{|c|}{Label Distribution Shift} & Original & Corruption-1 & Corruption-2 & Corruption-3 & Corruption-4 & Corruption-5\\\hline

    & Source & 95.0 ± 2.4 & 89.2 ± 9.9 & 83.9 ± 12.5 & 78.5 ± 16.5 & 71.2 ± 20.5 & 59.2 ± 26.5 \\\cdashline{2-8}[2pt/2pt]
    & TTN & -75.1 ± 3.9 & -70.2 ± 9.7 & -65.5 ± 12.0 & -60.5 ± 16.0 & -53.7 ± 19.7 & -42.2 ± 25.9 \\
    1 class & TENT  & -75.2 ± 2.2 & -70.2 ± 2.6 & -65.5 ± 2.7 & -60.4 ± 2.8 & -53.6 ± 2.7 & -42.1 ± 2.5 \\
    & LAME  &+2.4 ± 0.9 & +8.2 ± 1.1 & +10.6 ± 17.2 & +13.6 ± 22.4 & +13.3 ± 33.9 & +7.8 ± 45.7\\
    & DELTA  &-75.1 ± 2.3 & -70.2 ± 2.5 & -65.5 ± 2.7 & -60.5 ± 2.8 & -53.7 ± 2.7 & -42.2 ± 2.6 \\
    & Hybrid-TTN & -8.0 ± 3.1 & -9.2 ± 5.7 & -7.4 ± 7.9 & -5.1 ± 10.8 & -1.7 ± 14.9 & +3.8 ± 22.1
    \\\hline 
     
    & Source  & 95.2 ± 2.1 & 87.9 ± 10.3 & 82.0 ± 12.6 & 75.9 ± 16.7 & 68.1 ± 20.2 & 55.4 ± 22.7 \\\cdashline{2-8}[2pt/2pt]
    & TTN  & -32.4 ± 4.3 & -28.1 ± 8.4 & -24.2 ± 10.5 & -19.4 ± 14.2 & -13.4 ± 17.3 & -3.5 ± 21.3 \\
    3 classes & TENT  & -31.8 ± 4.6 & -27.4 ± 6.0 & -23.5 ± 6.5 & -18.7 ± 7.0 & -12.7 ± 7.8 & -2.7 ± 7.6 \\
    & LAME  &-0.8 ± 2.0 & +0.6 ± 13.1 & -3.2 ± 22.0 & -8.5 ± 30.5 & -10.6 ± 34.2 & -11.0 ± 33.6 \\
    & DELTA  &-32.4 ± 4.6 & -28.1 ± 5.9 & -24.2 ± 6.4 & -19.4 ± 7.0 & -13.4 ± 7.9 & -3.5 ± 7.6 \\
    & Hybrid-TTN & -3.3 ± 1.0 & -0.9 ± 5.3 & +2.0 ± 7.4 & +5.4 ± 10.1 & +10.0 ± 13.1 & +16.9 ± 18.1 \\\hline
    
    & Source  & 94.9 ± 1.2 & 88.5 ± 8.8 & 82.6 ± 10.3 & 76.5 ± 14.0 & 68.7 ± 17.3 & 56.5 ± 20.7  \\\cdashline{2-8}[2pt/2pt]
    & TTN  & -15.2 ± 2.0 & -13.0 ± 5.7 & -9.6 ± 7.2 & -4.9 ± 10.5 & +0.1 ± 13.4 & +9.0 ± 19.1 \\
    5 classes & TENT  & -14.8 ± 2.8 & -12.3 ± 5.3 & -9.0 ± 6.2 & -4.2 ± 7.1 & +0.8 ± 9.1 & +9.8 ± 9.0\\
    & LAME  &-2.1 ± 2.5 & -3.8 ± 13.0 & -9.9 ± 23.9 & -18.3 ± 31.2 & -22.4 ± 31.5 & -22.3 ± 26.6 \\
    & DELTA  &-15.2 ± 2.8 & -13.0 ± 5.2 & -9.6 ± 6.3 & -4.9 ± 7.3 & +0.1 ± 9.1 & +9.0 ± 9.2 \\
    & Hybrid-TTN & -2.2 ± 1.0 & +0.1 ± 4.7 & +3.0 ± 6.0 & +7.3 ± 9.1 & +11.9 ± 12.0 & +19.1 ± 17.3 \\\hline
    
    & Source & 93.5 ± 0.9 & 86.6 ± 9.3 & 80.2 ± 11.1 & 73.8 ± 14.7 & 65.6 ± 17.9 & 53.1 ± 20.1  \\\cdashline{2-8}[2pt/2pt]
    & TTN & -1.6 ± 0.6 & +2.2 ± 5.5 & +6.3 ± 7.0 & +10.8 ± 9.9 & +16.4 ± 12.9 & +25.2 ± 17.7 \\
    10 classes & TENT  & -1.5 ± 1.3 & +2.4 ± 4.3 & +6.5 ± 5.2 & +11.1 ± 6.7 & +16.8 ± 8.9 & +25.6 ± 9.6 \\
    (Original) & LAME  & -4.4 ± 1.3 & -14.1 ± 21.8 & -25.9 ± 28.0 & -33.3 ± 28.6 & -36.7 ± 25.1 & -35.9 ± 11.6 \\
    & DELTA  &-1.6 ± 1.3 & +2.2 ± 4.4 & +6.3 ± 5.4 & +10.8 ± 6.9 & +16.4 ± 9.1 & +25.2 ± 9.8 \\
    & Hybrid-TTN & -1.0 ± 0.6 & +2.3 ± 5.0 & +6.1 ± 6.5 & +10.4 ± 9.3 & +15.6 ± 12.2 & +23.1 ± 17.0 \\\hline

    & Source & 94.2 ± 1.8 & 87.5 ± 9.3 & 80.9 ± 12.6 & 76.0 ± 14.8 & 65.3 ± 19.5 & 53.7 ± 22.2 \\\cdashline{2-8}[2pt/2pt]
    & TTN & -22.4 ± 6.8 & -18.6 ± 8.0 & -15.0 ± 12.1 & -11.5 ± 14.8 & -3.4 ± 15.8 & +6.5 ± 19.8 \\
    Dirichlet & TENT  & -22.3 ± 6.8 & -19.3 ± 8.3 & -14.2 ± 11.1 & -9.3 ± 14.7 & -2.7 ± 18.4 & +6.9 ± 22.5 \\
    ($\alpha = 0.01$) & LAME  & -1.5 ± 1.5 & -1.2 ± 7.0 & -8.2 ± 17.1 & -14.0 ± 21.3 & -19.2 ± 20.5 & -16.8 ± 16.4 \\
    & DELTA  & -22.9 ± 6.7 & -20.1 ± 8.3 & -15.0 ± 11.1 & -10.0 ± 14.7 & -3.5 ± 18.2 & +6.1 ± 22.5 \\
    & Hybrid-TTN & -2.6 ± 1.1 & +0.2 ± 4.3 & +3.6 ± 7.3 & +6.6 ± 9.7 & +13.3 ± 14.2 & +20.3 ± 18.1 \\\hline

    & Source & 94.5 ± 1.4 & 86.9 ± 10.0 & 81.3 ± 11.3 & 74.3 ± 15.5 & 67.4 ± 17.0 & 53.4 ± 21.9 \\\cdashline{2-8}[2pt/2pt]
    & TTN & -10.0 ± 3.6 & -7.0 ± 7.4 & -3.1 ± 7.8 & +1.5 ± 12.1 & +6.7 ± 14.1 & +16.6 ± 19.1 \\
    Dirichlet & TENT  & -9.9 ± 3.4 & -7.4 ± 7.2 & -1.9 ± 8.5 & +2.1 ± 11.5 & +7.4 ± 14.7 & +17.8 ± 18.8 \\
    ($\alpha = 0.5$) & LAME  & -3.6 ± 1.4 & -6.0 ± 7.5 & -18.6 ± 18.3 & -25.5 ± 19.8 & -28.2 ± 20.7 & -29.3 ± 17.7 \\
     & DELTA  & -10.5 ± 3.6 & -7.9 ± 7.2 & -2.4 ± 8.4 & +1.4 ± 11.3 & +6.8 ± 14.7 & +17.1 ± 18.7 \\
    & Hybrid-TTN & -1.8 ± 0.9 & +1.4 ± 4.8 & +4.8 ± 6.2 & +9.4 ± 10.5 & +14.2 ± 12.2 & +22.0 ± 18.0 \\\hline

    & Source & 94.6 ± 0.9 & 87.3 ± 9.5 & 80.9 ± 10.9 & 73.6 ± 15.0 & 65.7 ± 17.5 & 53.9 ± 20.0 \\\cdashline{2-8}[2pt/2pt]
    & TTN & -1.9 ± 0.8 & +1.9 ± 5.5 & +6.0 ± 6.7 & +11.5 ± 10.4 & +16.7 ± 12.6 & +24.8 ± 17.5 \\
    Dirichlet& TENT  & -1.7 ± 0.8 & +2.1 ± 5.5 & +6.2 ± 6.8 & +11.0 ± 10.0 & +16.8 ± 13.1 & +25.7 ± 18.0 \\
    ($\alpha = 100$) & LAME  & -5.5 ± 1.1 & -14.5 ± 15.1 & -26.1 ± 20.5 & -33.4 ± 19.5 & -37.8 ± 17.7 & -35.7 ± 17.7 \\
    & DELTA  & -1.9 ± 0.8 & +1.9 ± 5.4 & +5.9 ± 6.7 & +10.7 ± 9.8 & +16.4 ± 12.9 & +25.2 ± 17.9 \\
    & Hybrid-TTN & -1.3 ± 0.6 & +2.1 ± 4.7 & +5.9 ± 6.4 & +10.9 ± 9.6 & +16.0 ± 12.1 & +23.0 ± 16.8 \\\hline
    \end{tabular}
        \caption{CIFAR-10 evaluations on multiple label shifted distributions and covariate shifts (corruptions) with different degrees of label imbalance. We show the source model accuracy and the improvement (or degradation) as a delta accuracy. We observe that the proposed method provides benefits over source model when there is no covariate shift, while avoiding catastrophic failures and allowing benefits over source when there are label distribution shifts. }
        \label{tab:cifar_main}
\end{table*}

\begin{table*}[]
\small
    \centering
    \begin{tabular}{|c|c|c|c|c|c|c|c|}\hline
    \multicolumn{2}{|c}{} & \multicolumn{6}{|c|}{Covariate Shift} \\\hline
    \multicolumn{2}{|c}{} & \multicolumn{6}{|c|}{Accuracy (\% or $\Delta$\%)  } \\\hline
    \multicolumn{2}{|c|}{Label Distribution Shift} & Original & Corruption-1 & Corruption-2 & Corruption-3 & Corruption-4 & Corruption-5\\\hline
    & Source & 70.8 ± 15.1 & 55.6 ± 21.3 & 44.6 ± 22.8 & 34.4 ± 23.6 & 22.8 ± 22.5 & 14.0 ± 19.5 \\\cdashline{2-8}[2pt/2pt]
    & TTN & -70.0 ± 15.6 & -54.1 ± 21.4 & -43.3 ± 22.8 & -33.2 ± 23.6 & -21.9 ± 22.3 & -13.2 ± 19.3 \\
    1 class & TENT  & -70.0 ± 1.0 & -53.8 ± 1.5 & -43.0 ± 1.7 & -32.8 ± 1.5 & -21.8 ± 1.5 & -13.5 ± 1.2\\
    & LAME  & +29.2 ± 0.0 & +40.8 ± 19.7 & +46.2 ± 29.4 & +41.8 ± 43.1 & +29.9 ± 50.2 & +17.3 ± 46.7 \\
    & DELTA  & -70.0 ± 1.0 & -53.8 ± 1.6 & -43.1 ± 1.7 & -32.8 ± 1.5& -21.8 ± 1.5& -13.5 ± 1.3 \\
    & Hybrid-TTN & -22.4 ± 6.7 & -17.1 ± 11.4 & -15.3 ± 12.5 & -10.6 ± 13.0 & -6.2 ± 11.9 & -3.2 ± 11.1 \\\hline 
     
    & Source & 69.5 ± 5.9 & 51.9 ± 10.6 & 41.5 ± 12.8 & 32.2 ± 15.9 & 21.5 ± 15.8 & 13.1 ± 14.0 \\\cdashline{2-8}[2pt/2pt]
    & TTN & -41.3 ± 4.5 & -30.5 ± 7.9 & -24.2 ± 9.7 & -17.7 ± 11.5 & -10.1 ± 10.8 & -5.0 ± 9.9  \\
    3 classes & TENT  & -40.4 ± 3.5 & -30.4 ± 5.6 & -23.7 ± 5.9 & -17.2 ± 6.7 & -9.6 ± 7.1 & -4.7 ± 6.4\\
    & LAME  & +9.6 ± 26.1 & -9.5 ± 19.0 & -3.3 ± 14.3 & -0.6 ± 20.1 & -1.4 ± 17.8 & -2.8 ± 16.1 \\
    & DELTA  &-41.6 ± 2.9 & -30.9 ± 5.4 & -24.2 ± 5.8 & -17.7 ± 6.6 & -10.0 ± 6.8 & -5.0 ± 6.2 \\
    & Hybrid-TTN & -7.2 ± 2.0 & -3.3 ± 7.1 & -1.8 ± 7.7 & +0.2 ± 7.7 & +2.3 ± 7.3 & +3.7 ± 6.7  \\\hline
    
    & Source  & 71.1 ± 5.3 & 53.7 ± 9.9 & 42.8 ± 13.2 & 32.7 ± 16.2 & 21.4 ± 15.8 & 13.2 ± 13.5 \\\cdashline{2-8}[2pt/2pt]
    & TTN & -26.5 ± 4.4 & -19.3 ± 5.8 & -14.1 ± 8.4 & -8.9 ± 9.4 & -3.4 ± 8.7 & -0.0 ± 7.7  \\
    5 classes & TENT  & -25.8 ± 4.1 & -18.6 ± 7.0 & -13.0 ± 8.0 & -7.7 ± 10.2 & -2.6 ± 10.4 & +0.7 ± 9.6 \\
    & LAME  &-4.2 ± 38.6 & -30.0 ± 17.6 & -20.8 ± 14.2 & -13.3 ± 18.6 & -10.0 ± 10.0 & -7.7 ± 8.9\\
    & DELTA  & -26.6 ± 4.1 & -19.7 ± 7.0 & -14.0 ± 7.9 & -8.7 ± 10.0 & -3.3 ± 10.1 & +0.1 ± 9.3 \\
    & Hybrid-TTN & -5.2 ± 2.5 & -1.0 ± 4.7 & +0.4 ± 6.4 & +2.4 ± 7.0 & +4.5 ± 6.5 & +4.9 ± 6.0 \\\hline
    
    & Source & 69.8 ± 3.4 & 50.7 ± 8.1 & 40.2 ± 10.9 & 31.0 ± 14.0 & 20.7 ± 14.0 & 13.0 ± 12.2 \\\cdashline{2-8}[2pt/2pt]
    & TTN & -14.7 ± 0.7 & -6.2 ± 3.7 & -2.5 ± 5.6 & +0.9 ± 6.8 & +3.9 ± 6.7 & +5.0 ± 5.9 \\
    10 classes & TENT  &-13.8 ± 3.7 & -5.5 ± 5.9 & -1.5 ± 8.0 & +2.1 ± 11.3 & +4.9 ± 11.8 & 5.7 ± 11.6 \\
    & LAME  & -59.8 ± 0.0 & -41.3 ± 0.0 & -30.8 ± 1.8 & -23.8 ± 4.5 & -16.3 ± 5.0 & -10.3 ± 4.4 \\
    & DELTA  & -14.7 ± 3.7 & -6.7 ± 6.0 & -2.7 ± 8.1 & +1.0 ± 11.3 & +3.9 ± 11.5 & +4.9 ± 11.2  \\
    & Hybrid-TTN & -2.9 ± 0.8 & +2.6 ± 3.3 & +3.5 ± 5.0 & +5.2 ± 6.0 & +5.9 ± 5.7 & +5.6 ± 5.3 \\\hline

    & Source & 69.7 ± 1.6 & 52.6 ± 7.3 & 42.0 ± 10.3 & 33.1 ± 14.0 & 23.1 ± 14.3 & 14.9 ± 12.3 \\\cdashline{2-8}[2pt/2pt]
    & TTN & -0.6 ± 0.6 & +5.9 ± 3.5 & +8.6 ± 5.2 & +10.4 ± 6.6 & +11.5 ± 6.4 & +10.7 ± 6.9 \\
    500 classes & TENT  & -0.5 ± 2.3 & +6.1 ± 5.5 & +9.2 ± 8.2 & +11.0 ± 12.1 & +12.2 ± 13.8 & +11.4 ± 14.5\\
    & LAME  & -55.2 ± 28.0 & -51.9 ± 6.6 & -41.8 ± 0.2 & -32.8 ± 0.2 & -22.8 ± 0.2 & -14.7 ± 0.1 \\
    & DELTA  & -0.6 ± 2.1 & +5.7 ± 5.8 & +8.8 ± 8.3 & +10.5 ± 12.2 & +11.7 ± 13.8 & +10.8 ± 14.4  \\
    & Hybrid-TTN & -0.2 ± 0.4 & +4.3 ± 3.1 & +5.6 ± 4.7 & +7.0 ± 5.8 & +7.5 ± 5.6 & +7.1 ± 5.9  \\\hline

    & Source & 70.0 ± 2.2 & 53.0 ± 7.3 & 42.0 ± 10.7 & 33.2 ± 13.7 & 23.1 ± 14.4 & 14.7 ± 12.3 \\\cdashline{2-8}[2pt/2pt]
    & TTN  & -1.0 ± 0.9 & +5.3 ± 3.4 & +8.1 ± 5.0 & +10.0 ± 6.6 & +11.5 ± 6.3 & +10.7 ± 6.9 \\
    Dirichlet & TENT & -0.8 ± 0.9 & +6.0 ± 3.4 & +9.0 ± 5.2 & +10.8 ± 6.5 & +11.8 ± 7.0 & +10.6 ± 6.6\\
    ($\alpha = 0.01$) & LAME & -67.1 ± 10.7 & -51.9 ± 7.0 & -40.8 ± 9.9 & -32.2 ± 13.7 & -22.9 ± 14.1 & -14.5 ± 12.7 \\
     & DELTA & -0.9 ± 0.9 & +5.6 ± 3.3 & +8.4 ± 5.2 & +10.2 ± 6.5 & +11.2 ± 7.0 & +10.1 ± 6.5  \\
    & Hybrid-TTN & -0.5 ± 0.8 & +4.4 ± 3.0 & +5.8 ± 4.5 & +7.0 ± 6.2 & +8.0 ± 5.6 & +7.4 ± 5.9 \\\hline

 \end{tabular}
 \caption{ImageNet-1K evaluations on multiple label shifted distributions and covariate shifts (corruptions) with different degrees of label imbalance. We observe that the proposed method provides benefits over source model when there is covariate shift, while avoiding catastrophic failures when there are label distribution shifts. Note that we use a batch size of 500 images, and the highest amount of classes is limited by this number.}
    \label{tab:imagenet}
\end{table*}

\begin{table*}[]
\small
    \centering
    \begin{tabular}{|c|c|c|c|c|c|c|}\hline
    \multicolumn{2}{|c}{} & \multicolumn{5}{|c|}{Covariate Shift} \\\hline
    \multicolumn{2}{|c}{} & \multicolumn{5}{|c|}{Accuracy (\% or $\Delta$\%)  } \\\hline
    \multicolumn{2}{|c|}{Label Distribution Shift} & Original & Corruption-1 & Corruption-2 & Corruption-3 & Corruption-4\\\hline

    & Source & 76.5 ± 2.7 & 74.8 ± 1.3 & 70.8 ± 3.1 & 65.3 ± 3.6 & 60.4 ± 5.1 \\\cdashline{2-7}[2pt/2pt]
    & TTN & +5.6 ± 4.2 & +3.2 ± 3.2 & +3.5 ± 5.5 & +4.4 ± 6.7 & +5.0 ± 9.4 \\
    Original & TENT & +5.8 ± 4.2 & +3.8 ± 5.0 & 0.0 ± 6.1 & +3.5 ± 9.1 & +8.0 ± 4.3 \\
    (31:69) & LAME  & -7.8 ± 2.7 & -4.9 ± 2.3 & -4.4 ± 2.9 & +2.8 ± 5.0 & +9.1 ± 3.4 \\
    & DELTA & +5.6 ± 4.2 & +3.2 ± 4.3 & -0.4 ± 6.0 & +3.0 ± 8.4 & +8.1 ± 4.1   \\
    & Hybrid-TTN & +0.9 ± 4.3 & -2.2 ± 3.5 & +3.5 ± 2.8 & +6.7 ± 4.9 & +10.5 ± 5.6 \\\hline
     
    & Source & 65.8 ± 3.2 & 60.5 ± 3.0 & 59.8 ± 3.5 & 53.9 ± 5.9 & 48.0 ± 3.0 \\\cdashline{2-7}[2pt/2pt]
    & TTN & +6.4 ± 4.9 & +6.1 ± 4.2 & +4.5 ± 6.0 & +5.5 ± 2.5 & +11.9 ± 4.6 \\
    Balanced & TENT & +6.9 ± 4.8 & +8.1 ± 5.2 & +6.6 ± 5.9 & +9.0 ± 6.8 & +15.1 ± 5.3\\
    (50:50) & LAME & -15.8 ± 3.2 & -10.4 ± 2.4 & -8.5 ± 3.3 & -3.8 ± 5.1 & +3.5 ± 4.2  \\
    & DELTA  & +6.4 ± 4.9 & +7.1 ± 4.9 & +6.6 ± 5.7 & +9.0 ± 6.7 & +14.8 ± 5.2 \\
    & Hybrid-TTN  & +8.2 ± 3.9 & +9.1 ± 3.6 & +5.8 ± 5.4 & +8.8 ± 7.0 & +10.6 ± 5.0 \\\hline
    
    & Source  & 34.6 ± 4.9 & 23.1 ± 3.6 & 22.4 ± 5.2 & 24.2 ± 2.8 & 15.9 ± 4.7 \\\cdashline{2-7}[2pt/2pt]
    & TTN & -11.9 ± 7.7 & +3.4 ± 4.1 & +4.1 ± 4.3 & +2.8 ± 4.4 & +9.9 ± 6.1 \\
    Healthy livers & TENT& -11.9 ± 8.1 & +2.4 ± 6.2 & +0.8 ± 5.7 & +2.6 ± 4.2 & +11.4 ± 4.7 \\
    (100:0) & LAME & -34.0 ± 5.2 & -23.0 ± 5.0 & -25.6 ± 3.8 & -24.5 ± 3.7 & -15.4 ± 5.1  \\
    & DELTA & -11.9 ± 7.7 & +2.2 ± 6.1 & +0.6 ± 5.5 & +2.4 ± 4.4 & +10.8 ± 5.2 \\
    & Hybrid-TTN & +29.0 ± 3.1 & +41.1 ± 5.0 & +19.4 ± 5.5 & +6.0 ± 6.9 & +5.4 ± 7.0 \\\hline
    
    & Source & 96.6 ± 1.2 & 98.8 ± 1.3 & 94.8 ± 3.1 & 86.1 ± 2.6 & 79.9 ± 6.1 \\\cdashline{2-7}[2pt/2pt]
    & TTN & -24.0 ± 3.7 & -24.6 ± 2.2 & -22.6 ± 3.3 & -12.8 ± 3.7 & -5.8 ± 8.5 \\
    Fatty livers & TENT & -24.0 ± 2.9 & -25.6 ± 2.4 & -22.9 ± 3.3 & -10.6 ± 4.9 & -7.8 ± 3.4 \\
    (0:100) & LAME & +3.4 ± 1.2 & +0.4 ± 0.6 & +5.6 ± 2.4 & +15.5 ± 4.4 & +19.2 ± 4.0\\
    & DELTA & -24.0 ± 3.7 & -26.0 ± 3.0 & -23.0 ± 3.3 & -10.0 ± 4.8 & -7.6 ± 4.3 \\
    & Hybrid-TTN & -21.4 ± 4.1 & -25.0 ± 6.4 & -11.4 ± 4.7 & +2.1 ± 6.2 & +10.0 ± 8.8 \\\hline

\hline
    \end{tabular}
    \caption{Evaluation on ultrasound data. Binary classification, subjects without fatty liver or patients with fatty liver disease. Models were trained on one site and are evaluated on a publicly available dataset \citep{byra}. We show evaluations on targets with shifts in label distribution and added speckle noise. We observe that the proposed method is able to control the degradation in performance under severe label distribution shift.} 
        \label{tab:ultrasound_deltas}
\end{table*}

\clearpage
\subsection{Results by corruption type} \label{sec:app_corruptions}
Similarly to other works on Test-time Adaptation, we display here the results separated by corruption type. In the interest of space, we show the results for accuracy in CIFAR-10 and Imagenet-1K, considering severity level 5 for all corruptions.

\begin{table}[h]
\centering
\resizebox{\textwidth}{!}{%
\begin{tabular}{l|c|c|c|c|c|c|c|c|c}
\hline
 & \multicolumn{4}{c|}{\textbf{CIFAR-10}} &  \multicolumn{5}{c}{\textbf{Imagenet-1K}} \\ \cline{2-10} 
\multicolumn{1}{c|}{Corruption / \# of Classes} & \textbf{1} & \textbf{3} & \textbf{5} & \textbf{10} & \textbf{1} & \textbf{3} & \textbf{5} & \textbf{10} & \textbf{500} \\ \hline
\textbf{Brightness} & $83.5 \pm 5.3$ & $89.4 \pm 2.6$ & $89.9 \pm 1.5$ & $90.4 \pm 1.9$ & $36.4 \pm 12.3$ & $46.8 \pm 9.3$ & $50.6 \pm 5.6$ & $49.8 \pm 3.9$ & $54.4 \pm 2.9$ \\ 
\textbf{Contrast} & $73.8 \pm 7.4$ & $82.4 \pm 3.8$ & $84.4 \pm 2.0$ & $84.7 \pm 1.8$ & $2.4 \pm 2.6$ & $2.9 \pm 3.3$ & $4.5 \pm 1.1$ & $5.8 \pm 3.9$ & $6.1 \pm 1.1$ \\ 
\textbf{Defocus Blur} & $72.2 \pm 9.4$ & $82.8 \pm 2.8$ & $83.7 \pm 2.6$ & $86.1 \pm 2.2$ & $2.0 \pm 1.4$ & $4.5 \pm 0.9$ & $4.5 \pm 1.6$ & $5.2 \pm 3.2$ & $7.5 \pm 0.9$ \\ 
\textbf{Elastic Transform} & $58.1 \pm 7.7$ & $71.1 \pm 5.6$ & $73.6 \pm 1.1$ & $75.0 \pm 2.0$ & $7.6 \pm 8.9$ & $22.1 \pm 10.3$ & $22.6 \pm 7.6$ & $25.0 \pm 4.3$ & $31.4 \pm 1.3$ \\ 
\textbf{Fog} & $67.5 \pm 12.3$ & $75.2 \pm 3.9$ & $80.1 \pm 1.7$ & $80.6 \pm 2.7$ & $14.4 \pm 7.3$ & $25.1 \pm 10.5$ & $27.4 \pm 5.0$ & $29.2 \pm 2.1$ & $32.8 \pm 2.1$ \\ 
\textbf{Frost} & $65.7 \pm 11.5$ & $70.0 \pm 5.8$ & $74.4 \pm 2.1$ & $77.2 \pm 2.1$ & $9.6 \pm 3.6$ & $16.9 \pm 5.6$ & $18.8 \pm 6.6$ & $20.2 \pm 3.6$ & $23.2 \pm 1.4$ \\ 
\textbf{Gaussian Blur} & $68.0 \pm 14.1$ & $81.1 \pm 3.5$ & $81.8 \pm 2.5$ & $85.0 \pm 1.4$ & $2.4 \pm 3.6$ & $3.7 \pm 2.1$ & $3.7 \pm 1.5$ & $4.6 \pm 3.2$ & $6.4 \pm 1.5$ \\ 
\textbf{Gaussian Noise} & $45.4 \pm 27.9$ & $50.2 \pm 11.7$ & $58.5 \pm 5.4$ & $58.6 \pm 2.0$ & $1.6 \pm 1.7$ & $6.0 \pm 4.5$ & $6.4 \pm 3.2$ & $6.0 \pm 1.9$ & $8.3 \pm 0.9$ \\ 
\textbf{Glass Blur} & $47.3 \pm 17.4$ & $59.1 \pm 9.4$ & $64.7 \pm 2.6$ & $61.8 \pm 1.8$ & $1.2 \pm 2.7$ & $5.6 \pm 4.0$ & $4.4 \pm 2.2$ & $5.9 \pm 1.6$ & $7.5 \pm 1.2$ \\ 
\textbf{Impulse Noise} & $46.6 \pm 20.9$ & $53.9 \pm 12.8$ & $59.3 \pm 6.1$ & $57.8 \pm 1.0$ & $0.8 \pm 1.1$ & $4.5 \pm 3.8$ & $4.6 \pm 2.6$ & $5.4 \pm 1.4$ & $6.9 \pm 1.0$ \\ 
\textbf{JPEG Compression} & $52.9 \pm 11.0$ & $69.6 \pm 5.0$ & $70.6 \pm 2.8$ & $72.6 \pm 1.3$ & $18.8 \pm 7.6$ & $21.7 \pm 5.4$ & $26.2 \pm 2.0$ & $25.7 \pm 5.0$ & $27.1 \pm 1.1$ \\ 
\textbf{Motion Blur} & $73.1 \pm 8.0$ & $81.2 \pm 3.0$ & $81.8 \pm 1.8$ & $84.0 \pm 1.8$ & $8.8 \pm 8.3$ & $10.8 \pm 3.7$ & $11.0 \pm 5.1$ & $12.0 \pm 6.5$ & $17.6 \pm 2.3$ \\ 
\textbf{Pixelate} & $59.1 \pm 11.3$ & $72.3 \pm 7.1$ & $75.4 \pm 3.5$ & $74.3 \pm 1.2$ & $18.4 \pm 13.7$ & $25.9 \pm 4.5$ & $26.0 \pm 5.1$ & $26.8 \pm 6.1$ & $32.4 \pm 4.0$ \\ 
\textbf{Saturate} & $82.3 \pm 6.7$ & $89.8 \pm 3.4$ & $91.6 \pm 1.8$ & $90.6 \pm 2.3$ & $34.0 \pm 11.8$ & $43.2 \pm 7.3$ & $44.9 \pm 5.4$ & $46.0 \pm 2.6$ & $50.4 \pm 3.2$ \\ 
\textbf{Shot Noise} & $47.9 \pm 26.8$ & $51.1 \pm 11.0$ & $60.2 \pm 4.9$ & $61.0 \pm 3.1$ & $4.0 \pm 4.9$ & $6.3 \pm 4.7$ & $5.8 \pm 4.2$ & $5.6 \pm 1.1$ & $8.0 \pm 1.5$ \\ 
\textbf{Snow} & $66.5 \pm 8.3$ & $75.1 \pm 4.3$ & $76.1 \pm 2.1$ & $77.5 \pm 2.8$ & $12.4 \pm 9.9$ & $17.6 \pm 6.5$ & $21.7 \pm 5.9$ & $19.3 \pm 2.5$ & $24.1 \pm 2.4$ \\ 
\textbf{Spatter} & $69.9 \pm 15.4$ & $81.5 \pm 7.0$ & $81.4 \pm 3.6$ & $81.1 \pm 1.8$ & $12.8 \pm 7.3$ & $25.1 \pm 5.0$ & $25.6 \pm 5.5$ & $25.9 \pm 2.4$ & $30.0 \pm 3.1$ \\ 
\textbf{Speckle Noise} & $49.2 \pm 25.4$ & $52.7 \pm 10.0$ & $61.4 \pm 4.8$ & $61.7 \pm 4.6$ & $7.2 \pm 4.4$ & $13.6 \pm 4.3$ & $15.0 \pm 4.4$ & $14.3 \pm 2.9$ & $16.3 \pm 1.2$ \\ 
\textbf{Zoom Blur} & $68.8 \pm 15.2$ & $84.4 \pm 3.5$ & $86.2 \pm 1.4$ & $87.8 \pm 2.6$ & $10.8 \pm 7.4$ & $16.8 \pm 3.2$ & $19.7 \pm 3.4$ & $21.0 \pm 3.9$ & $27.2 \pm 1.6$ \\ \hline
\end{tabular}%
}
\caption{Performance of Hybrid-TTN on CIFAR-10 and ImageNet-1K datasets under various corruptions and multiple imbalance scenarios.}
\label{tab:corruptions}
\end{table}

\begin{table}[h]
\small
\centering
\begin{tabular}{l|c|c|c|c|c|c}
\hline
 \textbf{Only 1-class} & \multicolumn{3}{c|}{\textbf{CIFAR-10}} & \multicolumn{3}{c}{\textbf{ImageNet-1K}} \\ \hline
 \textbf{Corruption} & \textbf{Source} & \textbf{TTN} & \textbf{Hybrid-TTN} & \textbf{Source}& \textbf{TTN} & \textbf{Hybrid-TTN} \\ \hline
 \textbf{Brightness} & $91.6 \pm 3.7$ & $19.2 \pm 2.8$ & $83.5 \pm 5.3$ & $46.4 \pm 23.0$ &$ 1.6 \pm 2.6$ & $36.4 \pm 12.3$ \\ 
\textbf{Contrast} & $36.5 \pm 26.7$ & $18.6 \pm 1.8$ & $73.8 \pm 7.4$ & $1.2 \pm 1.8$ & $ 0.4 \pm 0.9$& $2.4 \pm 2.6$ \\ 
\textbf{Defocus Blur} & $44.2 \pm 23.4$ & $17.6 \pm 2.5$ & $72.2 \pm 9.4$ & $6.0 \pm 9.3$ & $ 0.4 \pm 0.9$ & $2.0 \pm 1.4$ \\ 
\textbf{Elastic Transform} & $69.0 \pm 12.3$ & $17.2 \pm 1.6$ & $58.1 \pm 7.7$ & $5.2 \pm 7.6$ &$ 0.0 \pm 0.0$ & $7.6 \pm 8.9$ \\ 
\textbf{Fog} & $62.6 \pm 9.4$ & $18.4 \pm 3.1$ & $67.5 \pm 12.3$ & $9.6 \pm 12.0$ &$ 0.8 \pm 1.1$& $14.4 \pm 7.3$ \\ 
\textbf{Frost} & $67.5 \pm 14.9$ & $17.4 \pm 0.9$ & $65.7 \pm 11.5$ & $8.4 \pm 7.0$ &$ 2.4 \pm 2.2$& $9.6 \pm 3.6$ \\ 
\textbf{Gaussian Blur} & $32.7 \pm 31.5$ & $17.8 \pm 2.0$ & $68.0 \pm 14.1$ & $3.2 \pm 5.0$ &$ 0.4 \pm 0.9$& $2.4 \pm 3.6$ \\ 
\textbf{Gaussian Noise} & $40.4 \pm 33.3$ & $14.6 \pm 2.5$ & $45.4 \pm 27.9$ & $0.0 \pm 0.0$ &$ 0.8 \pm 1.1$& $1.6 \pm 1.7$ \\ 
\textbf{Glass Blur} & $58.7 \pm 24.3$ & $14.7 \pm 2.3$ & $47.3 \pm 17.4$ & $2.4 \pm 5.4$ &$ 0.0 \pm 0.0$& $1.2 \pm 2.7$ \\ 
\textbf{Impulse Noise} & $47.1 \pm 27.6$ & $13.4 \pm 2.2$& $46.6 \pm 20.9$ & $0.4 \pm 0.9$ &$ 0.0 \pm 0.0$& $0.8 \pm 1.1$ \\ 
\textbf{JPEG Compression} & $70.2 \pm 11.0$ & $16.5 \pm 1.8$& $52.9 \pm 11.0$ & $37.6 \pm 22.3$ &$ 1.2 \pm 1.1$& $18.8 \pm 7.6$ \\ 
\textbf{Motion Blur} & $67.7 \pm 15.2$ & $18.5 \pm 2.1$& $73.1 \pm 8.0$ & $17.2 \pm 23.7$ &$ 0.4 \pm 0.9$& $8.8 \pm 8.3$ \\ 
\textbf{Pixelate} & $40.0 \pm 33.9$ & $17.2 \pm 2.6$& $59.1 \pm 11.3$ & $26.8 \pm 20.9$ &$ 0.8 \pm 1.1$& $18.4 \pm 13.7$ \\ 
\textbf{Saturate} & $90.7 \pm 3.9$ & $18.9 \pm 2.1$& $82.3 \pm 6.7$ & $52.8 \pm 18.1$ &$ 1.2 \pm 1.1$& $34.0 \pm 11.8$ \\ 
\textbf{Shot Noise} & $48.8 \pm 30.5$ & $14.8 \pm 2.0$& $47.9 \pm 26.8$ & $0.8 \pm 1.8$ &$ 0.0 \pm 0.0$& $4.0 \pm 4.9$ \\ 
\textbf{Snow} & $72.3 \pm 8.4$ & $17.3 \pm 2.2$& $66.5 \pm 8.3$ & $8.4 \pm 8.6$ &$ 0.8 \pm 1.1$& $12.4 \pm 9.9$ \\ 
\textbf{Spatter} & $78.4 \pm 21.5$ & $16.6 \pm 2.7$ & $69.9 \pm 15.4$ & $15.6 \pm 7.1$ &$ 2.4 \pm 0.9$& $12.8 \pm 7.3$ \\ 
\textbf{Speckle Noise} & $53.7 \pm 27.3$ & $16.0 \pm 2.2$& $49.2 \pm 25.4$ & $4.8 \pm 7.6$ &$ 0.4 \pm 0.9$& $7.2 \pm 4.4$ \\ 
\textbf{Zoom Blur} & $52.2 \pm 26.9$ & $17.9 \pm 2.2$& $68.8 \pm 15.2$ & $18.8 \pm 17.2$ &$ 0.4 \pm 0.9$& $10.8 \pm 7.4$ \\ \hline
\end{tabular}%
\caption{Comparison of Hybrid-TTN to source model an TTN on CIFAR-10 and ImageNet-1K datasets under various corruptions at the most imbalanced scenario, containing only one class on the target data.}
\label{tab:corruptions}
\end{table}

\end{document}